\documentclass[12pt,english,12pt, draftclsnofoot, onecolumn]{IEEEtran}
\usepackage[T1]{fontenc}
\usepackage[latin9]{inputenc}
\usepackage{color}
\usepackage{verbatim}
\usepackage{float}
\usepackage{units}
\usepackage{mathtools}
\usepackage{amsmath}
\usepackage{amssymb}
\usepackage{graphicx}
\usepackage{setspace}
\makeatletter

\providecommand{\tabularnewline}{\\}
\floatstyle{ruled}
\newfloat{algorithm}{tbp}{loa}
\providecommand{\algorithmname}{Algorithm}
\floatname{algorithm}{\protect\algorithmname}

\usepackage{eqnarray}
\usepackage{mathrsfs}
\usepackage{epsfig}
\usepackage[english]{babel}
\usepackage{subfigure}
\usepackage{epstopdf}
\usepackage{import}
\usepackage{color}
\usepackage{colortbl}\usepackage{cite}
\usepackage{algorithm}

\usepackage{bbm}
\usepackage{cases}
\usepackage[normalem]{ulem}
\usepackage{array}
\usepackage{times}
\usepackage{color}
\usepackage{amsfonts}
\usepackage{psfrag}
\usepackage{fancyhdr}
 \usepackage{algorithmic}
\allowdisplaybreaks

\usepackage{dsfont}

\usepackage{babel}

\makeatother

\usepackage{babel}
\begin{document}
\title{Vehicular Cooperative Perception Through Action Branching and Federated
Reinforcement Learning\thanks{An extended abstract version of this work has been presented in IEEE
Asilomar 2020 \cite{AbdelazizAsilomar} and a conference version has
been submitted to IEEE ICC 2021 \cite{Abde2106:V2V}.}}
\author{\IEEEauthorblockN{Mohamed~K.~Abdel-Aziz,~\IEEEmembership{Student Member,~IEEE},
Cristina~Perfecto,~\IEEEmembership{Member,~IEEE}, Sumudu~Samarakoon,~\IEEEmembership{Member,~IEEE},
Mehdi~Bennis,~\IEEEmembership{Fellow,~IEEE}, and Walid~Saad,~\IEEEmembership{Fellow,~IEEE}}\thanks{M. K. Abdel-Aziz, S. Samarakoon, and M. Bennis are with the Centre
for Wireless Communications, University of Oulu, 90014 Oulu, Finland
(e-mails: mohamed.abdelaziz@oulu.fi; sumudu.samarakoon@oulu.fi; mehdi.bennis@oulu.fi).}\thanks{C.~Perfecto is with University of the Basque Country UPV/EHU, Spain
(email: cristina.perfecto@ehu.eus).}\thanks{W. Saad is with the Department of Electrical and Computer Engineering,
Virginia Tech, Blacksburg, VA 24061, USA (e-mail: walids@vt.edu).}}
\maketitle
\begin{abstract}
Cooperative perception plays a vital role in extending a vehicle's
sensing range beyond its line-of-sight. However, exchanging raw sensory
data under limited communication resources is infeasible. Towards
enabling an efficient cooperative perception, vehicles need to address
the following fundamental question: What sensory data needs to be
shared? at which resolution? and with which vehicles? To answer this
question, in this paper, a novel framework is proposed to allow reinforcement
learning (RL)-based vehicular association, resource block (RB) allocation,
and content selection of cooperative perception messages (CPMs) by
utilizing a quadtree-based point cloud compression mechanism. Furthermore,
a federated RL approach is introduced in order to speed up the training
process across vehicles. Simulation results show the ability of the
RL agents to efficiently learn the vehicles' association, RB allocation,
and message content selection while maximizing vehicles' satisfaction
in terms of the received sensory information. The results also show
that federated RL improves the training process, where better policies
can be achieved within the same amount of time compared to the non-federated
approach.
\end{abstract}

\begin{IEEEkeywords}
Cooperative perception, quadtree decomposition, federated reinforcement
learning, vehicle-to-vehicle (V2V) communication, association and
resource-block (RB) allocation.
\end{IEEEkeywords}

\section{Introduction\label{sec:Introduction}}

Recently, vehicles are rapidly becoming equipped with an increasing
variety of sensors (e.g., RADARs, LiDARs, and cameras) whose quality
varies widely \cite{WS1}. These sensors enable a wide range of applications
that assist and enhance the driving experience, from simple forward
collision and lane change warnings, to more advanced applications
of fully automated driving such as those of Waymo\footnote{www.waymo.com}(Google's
self-driving vehicles). Built-in sensors on these and other future
self-driving vehicles play a crucial role in autonomous navigation
and path planning. However, the reliability of these sensory information
is susceptible to weather conditions, existence of many blind spots
due to high density traffic or buildings, as well as sensors' manufacturing,
deployment, and operating defects, all of which may jeopardize the
success of these highly anticipated applications.

In order to overcome this issue, recent advancements in vehicle-to-vehicle
(V2V) communications (particularly as envisioned in future wireless
systems) can be utilized. V2V communications are seen as a promising
facilitator for intelligent transportation systems (ITS) \cite{extremeURLLC}.
It can ease the exchange of sensory information between vehicles to
enhance the perception of the surrounding environment beyond their
sensing range; such process is called \emph{cooperative perception
}\cite{etsi2019,MAVEN,machines5010006}. 
The advantages of cooperative perception are validated in \cite{Wang2018}, and \cite{DemonstrateCoperativePerception}, demonstrating its safety benefits and showing that it greatly improves the sensing performance. 
Motivated by its potential, several standardization bodies are currently focusing their efforts towards
formally defining the cooperative perception message (CPM), its contents
and generation rate \cite{3GPP,etsi2019,V2XStandardization}. In addition,
a growing body of literature has explored the use of cooperative perception
in various scenarios \cite{CPMGeneration,Gabb2019,WS3Zeng2019,Chen2019,3DObjectCooperativePerception,MLEnabledCooperativePerception}.
In \cite{CPMGeneration}, the authors investigated which information
should be included within the CPMs to enhance a vehicle's perception
reliability. Cooperative perception from the sensor fusion point-of-view
is studied in \cite{Gabb2019} and \cite{3DObjectCooperativePerception}. 
In \cite{Gabb2019}, a hybrid vehicular perception system
that fuses both local onboard sensor data as well as global received sensor data is proposed. Meanwhile in \cite{3DObjectCooperativePerception}, two novel cooperative 3D object detection schemes are proposed, named late and early fusion, depending on whether the fusion happens after or before the object detection stage. 
Moreover, in \cite{WS3Zeng2019},
the authors study the role of perception in the design of control and communications for platoons. 
The authors of \cite{Chen2019}
conducted a study on raw-data level cooperative perception for enhancing
the detection ability of self-driving systems; whereby sensory data
collected by every vehicle from different positions and angles of
connected vehicles are fused. Finally, the authors in \cite{MLEnabledCooperativePerception} discuss the challenges and opportunities of a machine-learning-enabled cooperative perception approach.
While interesting, none of these
works performs an in-depth analysis of the impact of wireless connectivity.

Cooperative perception over wireless networks cannot rely on exchanging
raw sensory data or point clouds, due to the limited communication
resources availability \cite{etsi2019}. For instance, a typical commercial
LiDAR using $64$ laser diodes produces $2.8$ million data points
per second with a horizontal and vertical field of views of $360^{\circ}$
and $26.8^{\circ}$ respectively, and a coverage range beyond $70\,\text{m}$.
Sharing even a small fraction of this information requires massive
data rates, which is why the use of millimeter wave (mmWave) communications
has been investigated in \cite{CristinaBeyondWYSIWYG} and \cite{V2XmmWave},
to leverage their high data rates and, thus, deal
with massive raw sensory data transmission. In order
to relax the data rate requirements and, hence, circumvent the use
of communications in the millimeter wave range, this raw sensory
data should be efficiently compressed to save both storage and available
communication resources. One possible technique that is useful for
such spatial raw sensory data is referred to as \emph{region quadtree}
\cite{QuadtreeHanan}. Region quadtree is a tree data structure used
to efficiently store data on a two-dimensional space. A quadtree recursively
decomposes the two-dimensional space into four equal sub-regions (blocks)
until all the locations within a block have the same state or until
reaching a maximum predefined resolution (tree-depth). Only a handful
of previous works, such as \cite{CarSpeak} and \cite{VehicularQuadtree2},
have used the quadtree concept within the vehicular networks domain.
In \cite{CarSpeak}, the authors introduced a communication system
for autonomous driving where a vehicle can query and access sensory
information captured by others. They used an octree, the 3D version
of quadtree, to model the world in order to allow vehicles to find
and query road regions easily. The authors in \cite{VehicularQuadtree2}
used the quadtree decomposition to find the minimal cost to relay
a message to a specific vehicle in a given geographical area. Within our work, the quadtree concept is utilized to model the sensory information
in the cooperative perception scenario. By doing so, a quadtree block
represents one of three states, either occupied, unoccupied or unknown,
and as a result, a vehicle could transmit specific quadtree blocks
covering a certain region instead of transmitting the corresponding
huge point cloud. Nonetheless, tailoring the number and resolution
of the transmitted quadtree blocks to bandwidth availability is a
challenging problem.

Moreover, simply broadcasting these sensory information (quadtree
blocks) to all neighboring vehicles, as suggested by \cite{etsi2019},
would impose a significant load on the available communication resources,
especially if the vehicular network is congested. Previous works have
tackled this problem in two ways: by filtering the number of objects
in the CPM to adjust the network load, as in \cite{CPMFiltering},
or by tweaking the generation rules of CPMs, as in \cite{CPMGeneration}
and \cite{aoki2020cooperative}. However, all these works still broadcast
the sensory information. Therefore, in order to mitigate the negative
effect of broadcasting, a principled approach to select which vehicles
should receive the relevant information, in which resolution and over
which resource blocks (RBs) is desperately needed, however,
complex.

Deep reinforcement learning (DRL) has proved useful in similar complex situations within the vehicular and wireless communication domains \cite{aoki2020cooperative,RL1,RL2,RL3Xianfu,RL4}. To the best of our knowledge, only \cite{aoki2020cooperative} lies within the cooperative perception scenario, where the main objective of DRL in a vehicular agent is to mitigate the network load by deciding either to transmit the CPM or discard it, without the ability to change its content. Although interesting, a proper selection of the content of the CPMs exchanged between the vehicles is needed to maximize the satisfaction of the vehicles with the received sensory information while complying with the available communication resources.

\subsection{Contributions}

The main contribution of this paper is a novel framework for solving
the joint problem of associating vehicles, allocating RBs, and selecting
the content of the CPMs exchanged between the vehicles,
with the objective of maximizing the mean satisfaction of all vehicles
with the received sensory information. Solving such a problem using
conventional mathematical tools is complex and intractable. As a result,
we resort to using machine learning techniques, specifically DRL \cite{Mnih2015}.

In particular, we split the main problem into two
sub-problems: The first problem focuses on associating vehicles and
allocating RBs, and solved at road-side unit (RSU) level, while the
other sub-problem focuses on selecting the content of the CPMs,
and is solved at the vehicle level. Both problems are formulated as
a DRL problem where the objective of the RSU is to learn the association
and RB allocation that yields a higher average vehicular satisfaction,
while the objective of each vehicle is to learn which sensory information
is useful and should be transmitted to its associated vehicle. Moreover,
in order to enhance the training process, we propose the use of federated
RL \cite{FRL,FLOpenProblems,EdgeML}. %
Specifically, at every time frame, each vehicle under the coverage
of the RSU shares its latest model parameters with the RSU, the RSU
then averages all the received model parameters and broadcasts the
outcome back to the vehicles under its coverage. Simulation results
show that the policies achieving higher vehicular satisfaction could
be learned at both the RSU and vehicles level. Moreover, the results
also show that federated RL improves the training process, where better
policies can be achieved within the same amount of time compared to
non-federated approach. Finally, it is shown that trained agents always
outperform non-trained random agents in terms of the achieved vehicular
satisfaction.

In a nutshell, the main contributions of this work can be summarized
as follows: \begin{itemize} \item We mathematically formulate the
joint problem of vehicle association, RB allocation and content selection
of the CPMs while taking into consideration the impact of the wireless
communication bandwidth.\item We propose an RL problem formulation
for vehicle association and RB allocation, as well as the RL problem
of the content selection of the CPMs. Moreover, to overcome the huge
action space inherent to the formulation of the RL problems, we apply
the dueling and branching concepts proposed in \cite{ActionBranching}.\item
We propose a federated RL approach to enhance the training process
of all vehicles.\item We conduct simulations based on practical traffic
data to demonstrate the effectiveness of the proposed approaches.\end{itemize}

The rest of this paper is organized as follows. In Section \ref{sec:System-model},
the different parts of the system model are described, including the
sensory, wireless communication, and quadtree models. The network-wide
problem is formulated in Section \ref{sec:Problem-formulation}, followed
by a brief introduction to RL and how it is utilized within our cooperative
perception scenario, in Section \ref{sec:Reinforcement-learning}.
In Section \ref{sec:Overcoming-the-huge}, the huge action space issue
and how to overcome it, is presented. The federated RL approach is
described in Section \ref{sec:Federated-RL}. Finally, in Section
\ref{sec:Numerical-results}, simulation results are presented while
conclusions are drawn in Section \ref{sec:Conclusion}.

\section{System Model\label{sec:System-model}}

\begin{figure}[t]
\centering \includegraphics[width=0.9\textwidth]{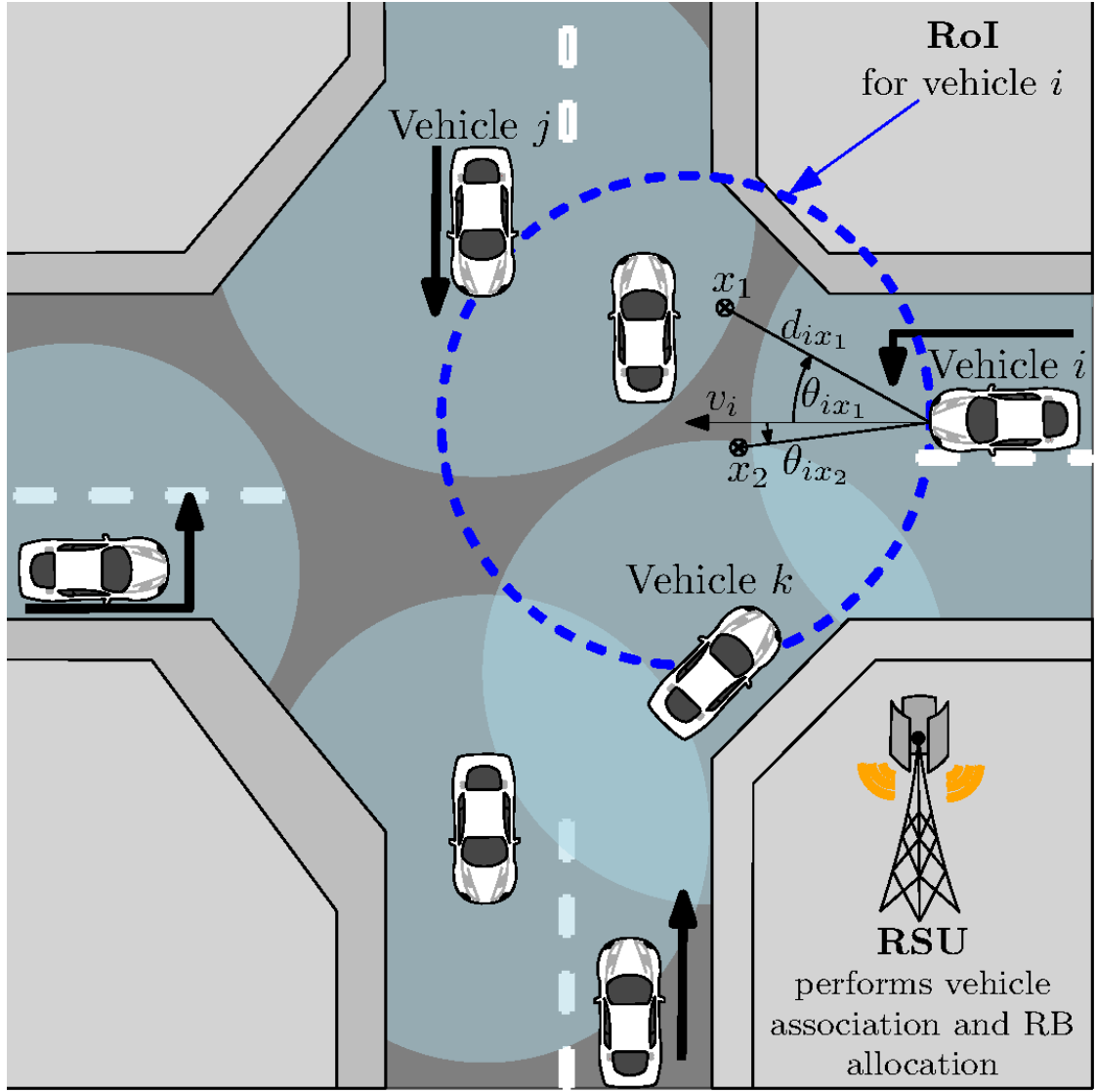}
\caption{Vehicles under the coverage of a single RSU, drive through a junction
while dynamically exchanging sensory information.}
\label{fig:1}
\end{figure}
Consider a road junction covered and serviced by a single RSU, as
shown in Fig.\,\ref{fig:1}. We consider a time-slotted
system with a slot index $t$ and a slot duration of $\tau$. Let
$\mathcal{N}_t$ be the set of $N$ vehicles
served by the RSU at time slot $t$, where $N$ is the maximum number of vehicles that could be served simultaneously by the RSU. We denote the location of each vehicle $n\in\mathcal{N}_t$
at time slot $t$ by $\boldsymbol{l}_{n}\left(t\right)$ and assume
that each vehicle is equipped with a sensor having a fixed circular
range of radius $r$. From a vehicle's perspective, the output of the sensor regarding any
location can have one of three states: Occupied ($s_{+}$), unoccupied
($s_{-}$), or unknown ($s_{0}$). The occupied state $s_+$ corresponds to locations within the vehicle's sensing range where an obstacle is sensed. While, the unoccupied state $s_-$ corresponds to locations within the vehicle's sensing range that are sensed to be obstacle-free. Finally, the unknown state $s_0$ corresponds to the locations which cannot be sensed by the vehicle's sensor, either due to occlusions or because they are outside the vehicle's sensing range. Lets assume that the output of vehicle $n$'s sensor regarding location $\boldsymbol{x}$ at time slot $t$ is $s_{n}(\boldsymbol{x},t)\in\{s_+,s_-,s_0\}$. Moreover, here the ``ground-truth'' state of any location $\boldsymbol{x}$ is either occupied $s_{+}$, or unoccupied $s_{-}$, and is given by $G(\boldsymbol{x},t)\in\{s_+,s_-\}$. Since sensors are not perfectly reliable, then $s_{n}(\boldsymbol{x},t)$ will be correct, i.e., $s_{n}(\boldsymbol{x},t)=G(\boldsymbol{x},t)$, with a fixed probability of $\lambda_n$, where $\lambda_{n}\in(0.5,1]$ represents the sensor's reliability. Thus, the probability of occupancy at location $\boldsymbol{x}$
with respect to vehicle $n$, given its sensor output, is, 
\begin{equation}
p_{n}(\boldsymbol{x},t)=\text{Pr}(G(\boldsymbol{x},t)=s_+|s_{n}(\boldsymbol{x},t))=\begin{cases}
\lambda_{n} & \text{if~}s_{n}(\boldsymbol{x},t)=s_{+},\\
1-\lambda_{n} & \text{if~}s_{n}(\boldsymbol{x},t)=s_{-},\\
1/2 & \text{if~}s_{n}(\boldsymbol{x},t)=s_{0},
\end{cases}\label{eq:OccupancyProbabilityPoint}
\end{equation}
Note that, some locations are occluded by obstacles, or are outside the sensing range, which results in $s_{n}(\boldsymbol{x},t)=s_{0}$.
In order to model the high uncertainty of this unknown state, it is
mapped to $p_{n}(\boldsymbol{x},t)=\frac{1}{2}$ which represents
the highest uncertainty in a probability distribution.

Let $q_{n}(\boldsymbol{x},t)$ be the value (or quality)
of the sensed information at location $\boldsymbol{x}$ at the beginning
of time slot $t.$ $q_{n}(\boldsymbol{x},t)$ depends on the probability
of occupancy $p_{n}(\boldsymbol{x},t)$ and the age of the information
(AoI)\footnote{AoI is defined as the time elapsed since the generation
instant of the sensory information.} $\Delta_{n}(\boldsymbol{x},t)$ \cite{AoI1,AoI2},
which is given by,
\begin{equation}
q_{n}(\boldsymbol{x},t)=|2p_{n}(\boldsymbol{x},t)-1|\mu^{\Delta_{n}(\boldsymbol{x},t)},\label{eq:InfoWorthinessPoint}
\end{equation}

\[
\Delta_{n}(\boldsymbol{x},t)=\tau t-\Gamma_{n}(\boldsymbol{x}),
\]
with a parameter $\mu\in(0,1)$ and $\Gamma_{n}(\boldsymbol{x})$
defining the instant when $\boldsymbol{x}$ was last sensed by vehicle
$n$.\footnote{Hereafter, the dependence on $t$ will be omitted
from the notations for the simplicity of presentation and brevity.} Here, we choose the AoI as a metric to emphasize the importance of
fresh sensory information. Note that the value function $q_{n}(\boldsymbol{x})$
decreases as its AoI increases (\emph{outdated} information) or the
probability of occupancy for location $\boldsymbol{x}$ approaches
1/2 (\emph{uncertain} information).\footnote{Note that, when location $\boldsymbol{x}$ is occupied, \eqref{eq:InfoWorthinessPoint} does not distinguish between the different kinds of objects. However, this can easily be mitigated by allowing $\mu$ to be inversely proportional to the speed of the detected object, e.g., $\mu_{\text{fast object}}<\mu_{\text{slow object}}<\mu_{\text{static object}}=1$.}

Moreover, each vehicle $n$ is interested in extending
its sensing range. The higher the vehicle's velocity is, the bigger
this region of interest (RoI) should be. As a result, each vehicle
is interested in $t_{\text{int}}$ seconds ahead along its direction
of movement. The RoI of vehicle $n$, for simplicity, is captured
by a circular region with a diameter of $v_{n}t_{\text{int}}$, where
$v_{n}$ is the velocity of the vehicle. Within the RoI, the vehicle
has higher interest in gaining sensory information regarding the locations
closer to its current position as well as regarding locations closer
to its direction of movement. Therefore, we formally define the interest
of vehicle $n$ in location $\boldsymbol{x}$ as follows: 
\begin{equation}
w_{n}(\boldsymbol{x})=\begin{cases}
\frac{v_{n}t_{\text{int}}\cos\theta-d}{v_{n}t_{\text{int}}\cos\theta}, & d\leq v_{n}t_{\text{int}}\cos\theta,\\
0, & \text{o.w.},
\end{cases},
\end{equation}
where $d$ is the euclidean distance between the location $\boldsymbol{x}$
and the vehicle's position $\boldsymbol{l}_{n}\left(t\right)$, and
$\theta$ is the angle between the vehicle's direction of motion and
location $\boldsymbol{x}$, as illustrated in Fig.\,\ref{fig:1}.
To capture the need of gathering new information, the interest $w_{n}(\boldsymbol{x})$
of vehicle $n$ needs to be weighted based on the lack of worthy information,
i.e., $1-q_{n}(\boldsymbol{x})$. Hence, the modified interest of
vehicle $n$ in location $\boldsymbol{x}$ is given by, 
\begin{equation}
i_{n}(\boldsymbol{x})=w_{n}(\boldsymbol{x})[1-q_{n}(\boldsymbol{x})].\label{eq:RoIWeights}
\end{equation}

We assume that each vehicle can associate with at
most one vehicle at each time slot to exchange sensory information.
We define $E(t)=\left[e_{nn'}(t)\right]$ to be the global association
matrix,where $e_{nn'}(t)=1$ if vehicle $n$ is associated (transmits)
to vehicle $n'$ at time slot $t$, otherwise, $e_{nn'}(t)=0$. It
is assumed that the association is bi-directional, i.e., $e_{nn'}(t)=e_{n'n}(t)$.
Moreover, we assume that each associated pair can communicate simultaneously
with each other, i.e. each vehicle is equipped with two radios, one
for transmitting and the other is for receiving. Additionally, a set
$\mathcal{K}$ of $K$ orthogonal resource blocks (RBs), with bandwidth
$\omega$ per RB, is shared among the vehicles, where each transmitting
radio is allocated with only one RB. We further define $\eta_{nn'}^{k}(t)\in\left\{ 0,1\right\} $, $\text{for all }k\in\mathcal{K}\text{ and }n,n'\in\mathcal{N}$, as the RB allocation.
Here, $\eta_{nn'}^{k}(t)=1$ if vehicle $n$ transmits over RB $k$
to vehicle $n'$ at time slot $t$ and $\eta_{nn'}^{k}(t)=0$, otherwise.
In order to avoid self-interference, the RBs allocated
for each associated pair are orthogonal, i.e. $\eta_{nn'}^{k}(t)\ne\eta_{n'n}^{k}(t)$
given that $e_{nn'}(t)=1$.

Let $h_{nn'}^{k}(t)$ be the instantaneous channel gain, including
path loss and channel fading, from vehicle $n$ to vehicle $n'$ over
RB $k$ in slot $t$. We consider the $5.9\text{ GHz}$ carrier frequency
and adopt the realistic V2V channel model of \cite{ChannelModel}
in which, depending on the location of the vehicles, the channel model
is categorized into three types: line-of-sight, weak-line-of-sight,
and non-line-of-sight. As a result, the data rate from vehicle $n$
to vehicle $n'$ at time slot $t$ (in packets per slot) is expressed
as 
\begin{equation}
R_{nn'}(t)=e_{nn'}(t)\cdot\frac{\tau}{M}\sum_{k\in\mathcal{K}}\eta_{nn'}^{k}(t)\omega\log_{2}\left(1+\frac{Ph_{nn'}^{k}(t)}{N_{0}\omega+I_{nn'}^{k}(t)}\right),\label{eq:rate}
\end{equation}
where $M$ is the packet length in bits, $P$ is the transmission
power per RB, and $N_{0}$ is the power spectral density of the additive
white Gaussian noise. Here, $I_{nn'}^{k}(t)=\sum_{i,j\in\mathcal{N}/n,n'}\eta_{i,j}^{k}(t)Ph_{in'}^{k}(t)$
indicates the received aggregate interference at the receiver $n'$
over RB $k$ from other vehicles transmitting over the same RB $k$.

\subsection{Quadtree Representation}

\begin{figure}[t]
\centering%
\begin{tabular}{cc}
\includegraphics[width=0.35\textwidth]{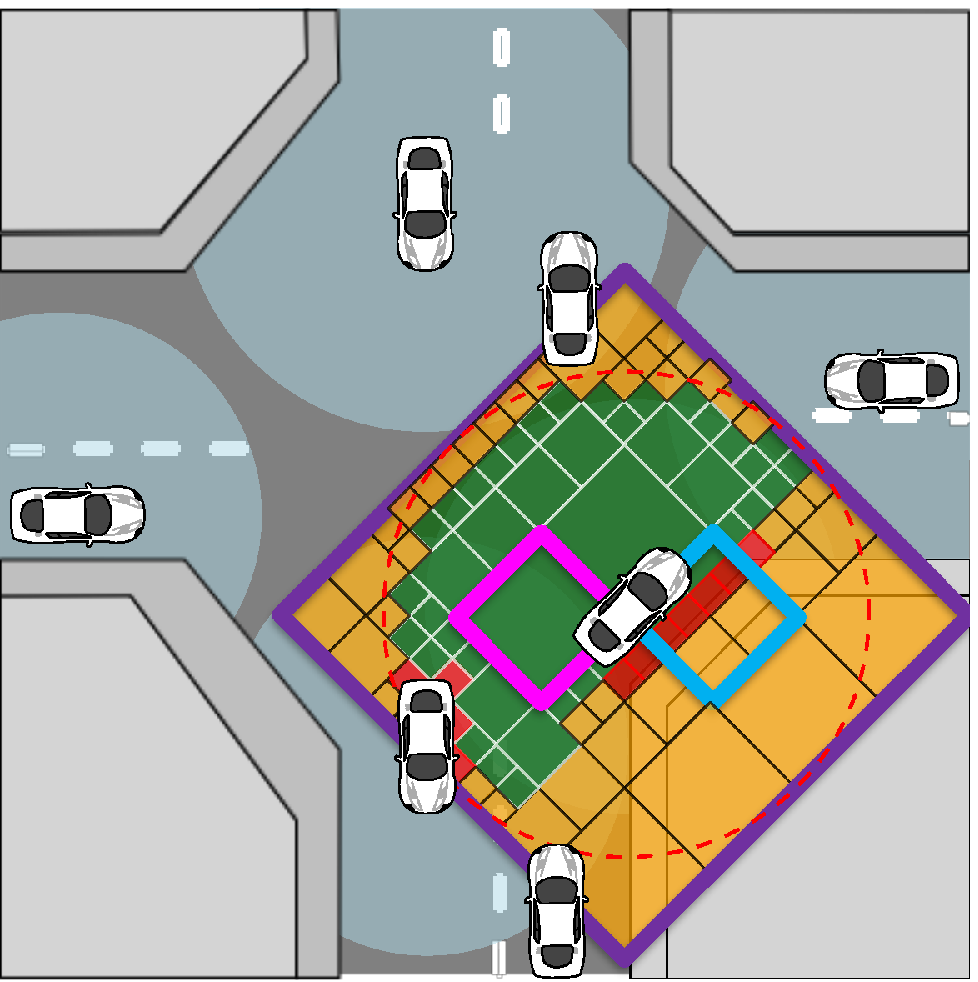} & \includegraphics[width=0.6\textwidth]{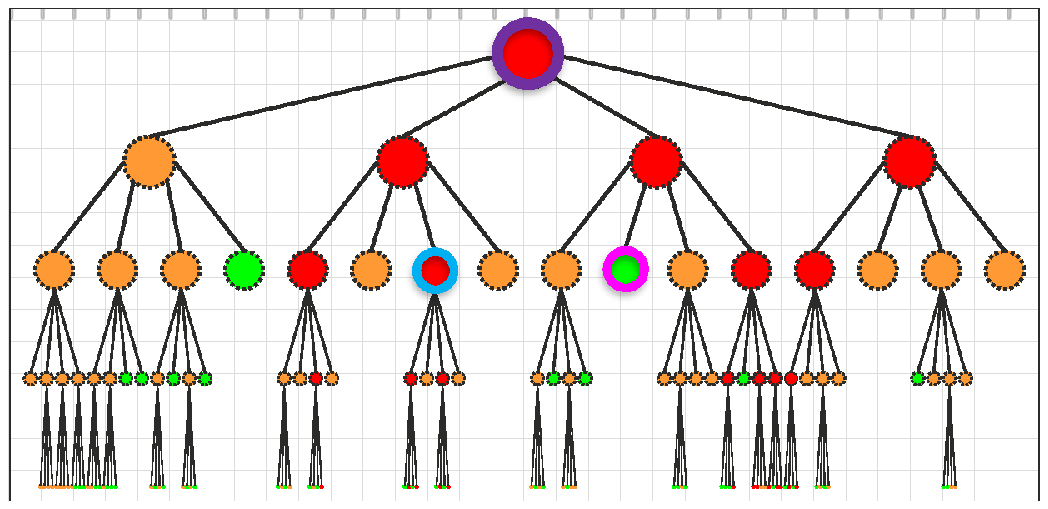}\tabularnewline
(a) & (b)\tabularnewline
\end{tabular} \caption{Quadtree representation of the sensing range of vehicle $k$, with
a maximum resolution level $L=5$. Green represents the unoccupied
state $s_{-}$, red represents the occupied state $s_{+}$ and orange
represents the unknown state $s_{0}$. (a) shows the block decomposition
of the sensing range while (b) shows the equivalent tree diagram.
The block with purple boarders represents the lowest
resolution block that can be transmitted. Another two examples of
different blocks are highlighted in the same way using pink and blue
boarders.}
\label{fig:2}
\end{figure}

Storing and exchanging raw sensory information between vehicles, e.g.,
information about individual locations $\boldsymbol{x}$, requires
significant memory and communication resources for cooperative perception
to be deemed useful. To alleviate this challenge, a compression technique
called \emph{region quadtree}, which efficiently store data on a two-dimensional
space, can be used by each vehicle \cite{QuadtreeHanan}. In this
technique, each vehicle converts its sensing range into a squared-block
of side-length $2r$. This block is divided recursively into 4 blocks
until 
\begin{itemize}
\item reaching a maximum resolution level $L$, or 
\item the state of every location $\boldsymbol{x}$ within a block is the same.
\end{itemize} 
Without loss of generality, we assume that each block can be represented using
$M$ bits. Fig.\,\ref{fig:2} shows the quadtree representation of
the sensing range of vehicle $k$ with $L=5$.

The state of block $b$ within the quadtree of vehicle $n$ is either,
\begin{itemize}
\item Occupied: If the state of any location $\boldsymbol{x}$ within
the block is occupied,
\item Unoccupied: If every location within the block is unoccupied, 
\item Unknown: Otherwise.
\end{itemize}
In this view, the probability of occupancy
of each block $p_{n}(b)$ can be defined in the same manner as \eqref{eq:OccupancyProbabilityPoint}:

\begin{equation}
p_{n}(b)=\begin{cases}
\lambda_{n} & \text{if~}s_{n}(b)=s_{+},\\
1-\lambda_{n} & \text{if~}s_{n}(b)=s_{-},\\
1/2 & \text{if~}s_{n}(b)=s_{0},
\end{cases}\label{eq:OccupancyProbabilityBlock}
\end{equation}
and the worthiness of block $b$'s sensory information $q_{n}(b)$
is defined in the same manner as \eqref{eq:InfoWorthinessPoint}.

Let $\mathcal{B}_{n}\left(t\right)$ represent the set of quadtree
blocks available for transmission by vehicle $n$ at the
beginning of time slot $t$. Assume that $\mathcal{B}_{n}\left(t\right)=\mathcal{B}_{n}^{\text{c}}\cup\mathcal{B}_{n}^{\text{p}}$,
where $\mathcal{B}_{n}^{\text{c}}$ is the set of blocks available
from its own current sensing range, while $\mathcal{B}_{n}^{\text{p}}$
is the set of blocks available from previous slots (either older own
blocks or blocks received from other vehicles). Note that, due to
the quadtree compression, the cardinality of $\mathcal{B}_{n}^{\text{c}}$
is upper bounded by: $|\mathcal{B}_{n}^{\text{c}}|\leq\sum_{l=0}^{L-1}4^{l}=\frac{1-4^{L}}{1-4}$.
Also, in order to keep the exchanged sensory information fresh, an
upper bound is applied on the cardinality of $\mathcal{B}_{n}^{\text{p}}$:
$|\mathcal{B}_{n}^{\text{p}}|\leq B_{\text{max}}^{\text{p}}$, where
blocks with higher AoI are discarded if the cardinality of $\mathcal{B}_{n}^{\text{p}}$
exceeded $B_{\text{max}}^{\text{p}}$. Determining what quadtree blocks
needs to be shared, and with which vehicles, is not straightforward.
In order to answer those questions, we first start by formulating
the problem.

\section{Problem Formulation \label{sec:Problem-formulation}}

In our model, each vehicle $n$ is interested in associating (pairing)
with another vehicle $n'$ where each pair exchanges sensory information
in the form of quadtree blocks with the objective of maximizing the
joint satisfaction of both vehicles. The satisfaction of vehicle $n$
with the sensory information received from vehicle $n'$ at time slot
$t$ can be defined as follows:

\begin{equation}
f_{nn'}\left(t\right)=\sum_{b\in\mathcal{B}_{n'}\left(t\right)}\sigma_{n'}^{b}\left(t\right)\left(\frac{\sum_{\boldsymbol{x}\in b}i_{n}\left(\boldsymbol{x}\right)}{\Lambda\left(b\right)}.q_{n'}\left(b\right)\right),\label{eq:VehicularSatisfaction}
\end{equation}
where $\sigma_{n'}^{b}\left(t\right)=1$ if vehicle $n'$ transmitted
block $b$ to vehicle $n$ at time slot $t$, and $\sigma_{n'}^{b}\left(t\right)=0$
otherwise, and $\Lambda\left(b\right)$ is the area covered by block
$b$. Moreover, it should be noted that vehicle $n$ is more satisfied
with receiving quadtree blocks with a resolution proportional to the
weights of its RoI as per \eqref{eq:RoIWeights}, i.e., block $b$
with higher resolution (smaller coverage area $\Lambda\left(b\right)$)
for the regions with higher $i_{n}\left(\boldsymbol{x}\right)$, which
is captured by $\frac{\sum_{\boldsymbol{x}\in b}i_{n}\left(\boldsymbol{x}\right)}{\Lambda\left(b\right)}$.
Furthermore, vehicle $n$ is more satisfied with receiving quadtree
blocks having more worthy sensory information, which is captured by
$q_{n'}\left(b\right)$. %
As a result, our cooperative perception network-wide problem can be
formally posed as follows:\begin{subequations}\label{First_optimization_problem}
\begin{align}
\max_{\boldsymbol{\eta}(t),E(t),\boldsymbol{\sigma}\left(t\right)} & ~~\sum\limits _{n,n'\in\mathcal{N}}f_{nn'}\left(t\right)\cdot f_{n'n}\left(t\right)\nonumber \\
\text{s.t.} & ~~\sum_{b\in\mathcal{B}_{n}\left(t\right)}\sigma_{n}^{b}\left(t\right)\leq\sum_{n'\in\mathcal{N}}R_{nn'}(t),\ \forall n\in\mathcal{N},\,\forall t,\label{eq:TransmittedBlocksBound}\\
 & ~~{\textstyle \sum\limits _{n'\in\mathcal{N}}}\sum_{k\in\mathcal{K}}\eta_{nn'}^{k}(t)\leq1,~\forall n\in\mathcal{N},\,\forall t,\label{eq:RBConstraint}\\
 & ~~\sum_{n'\in\mathcal{N}}e_{nn'}\left(t\right)\leq1,\ \forall n\in\mathcal{N},\,\forall t,\label{eq:AssociationConst1}\\
 & \ \ e_{nn'}\left(t\right)=e_{n'n}\left(t\right),\ \forall n,n'\in\mathcal{N},\,\forall t,\label{eq:AssociationConst2}\\
 & ~~\eta_{nn'}^{k}(t)\in\{0,1\},e_{nn'}\left(t\right)\in\left\{ 0,1\right\} ,\sigma_{n}^{b}\left(t\right)=\left\{ 0,1\right\} ~\forall t,\,k\in\mathcal{K},\,n,n'\in\mathcal{N},\label{eq:OptVariables}
\end{align}
\end{subequations}where the objective is to associate vehicles $E\left(t\right)$,
allocate RBs $\boldsymbol{\eta}\left(t\right)=[\eta_{nn'}^{k}(t)]^{k\in\mathcal{K}}_{n,n'\in\mathcal{N}}$, and select the contents
of the transmitted messages (the quadtree blocks to be transmitted
by each vehicle) $\boldsymbol{\sigma}\left(t\right)$, in order to
maximize the sum of the joint satisfaction of the associated vehicular
pairs. Note that \eqref{eq:TransmittedBlocksBound} is an upper bound
on the number of transmitted quadtree blocks of each vehicle by its
Shannon data rate, while \eqref{eq:RBConstraint} constrains the number
of RBs allocated to each vehicle to 1. 

Note that, the RB allocation $\eta_{nn'}^{k}(t)$ and the vehicular association $E\left(t\right)$ directly determine the data rate $R_{nn'}(t)$ as per \eqref{eq:rate}, specifying the maximum number of quadtree blocks to be transmitted between the vehicles. Moreover, the selected quadtree blocks $\boldsymbol{\sigma}\left(t\right)$ along with $E\left(t\right)$ directly determines the vehicular satisfaction as per \eqref{eq:VehicularSatisfaction}. As a result finding the optimal solution
(RB allocation, vehicular association and message content selection)
of this problem is complex and not straightforward. From a centralized
point of view where the RSU tries to solve this problem, the RSU needs
to know the real-time wireless channels between the vehicles and the
details of the sensed information of each vehicle, in order to optimally
solve \eqref{First_optimization_problem}. Frequently exchanging such
fast-varying information between the RSU and vehicles can yield a
huge communication overhead which is impractical. From a decentralized
point of view, in order to maximize \eqref{eq:VehicularSatisfaction},
vehicle $n'$ needs to know the exact interest of vehicle $n$ as
per \eqref{eq:RoIWeights} in order to optimally select the quadtree
blocks to be transmitted, which is impractical as well. Hence, to
solve \eqref{First_optimization_problem} we leverage machine learning
techniques which have proved to be useful in dealing with such complex
situations, specifically DRL \cite{Mnih2015}.

\section{Reinforcement Learning Based Cooperative Perception\label{sec:Reinforcement-learning}}

\subsection{Background}

RL is a computational approach to understanding goal-directed learning
and decision-making \cite{sutton2018reinforcement}. RL is about learning
from interactions how to behave in order to achieve a goal. The learner
(or decision-maker) is called an \emph{agent }who interacts with the
\emph{environment}, which is comprising everything outside the agent.

Thus, any goal-directed learning problem can be reduced to three signals
exchanged between an agent and its environment: one signal representing
the choices made by the agent (\emph{actions}), one signal representing
the basis on which the choices are made (\emph{states}), and one signal
defining the agent's goal (\emph{rewards}). In a typical RL problem,
the agent's goal is to maximize the total amount of reward it receives,
which means maximizing not just the immediate reward, but a cumulative
reward in the long run.

RL problems are typically formalized using Markov decision processes\footnote{Even when the state signal is not Markovian, it is still appropriate
to think of the state in reinforcement learning as an approximation
to a Markov state. \cite{sutton2018reinforcement}} (MDPs) \cite{sutton2018reinforcement}, characterized as $\left\langle \mathcal{S},\mathcal{A},\mathcal{T},\mathcal{R},\gamma\right\rangle $.
That is, at timestep $t$, the agent with state $s\in\mathcal{S}$
performs an action $a\in\mathcal{A}$ using a policy $\pi\left(a|s\right)$,
and receives a reward $r_{t}=\mathcal{R}\left(s,a\right)\in\mathbb{R}$,
and transitions to state $s'\in\mathcal{S}$ with probability $p\left(s'|s,a\right)=\mathcal{T}\left(s,a,s'\right)$.
We define $R_{t}=\sum_{t'=t}^{H}\gamma^{t'-t}r_{t}$ as the discounted
return over horizon $H$ and discount factor $\gamma\in\left[0,1\right)$,
and we define $Q^{\pi}\left(s,a\right)=\mathbb{E}_{\pi}\left[R_{t}|s_{t}=s,a_{t}=a\right]$
as the action-value (Q-value) of state $s$ and action $a$. Moreover,
let $\pi^{*}$ be the optimal policy that maximizes the Q-value function,
$Q^{\pi^{*}}\left(s,a\right)=\max_{\pi}Q^{\pi}\left(s,a\right)$.
The ultimate goal of RL is to learn the optimal policy $\pi^{*}$
by having agents interacting with the environment.

Among the various techniques used to solve RL problems, in this work
we will advocate for the use of Q-learning and deep Q-networks (DQNs)%
.

\subsubsection{Q-learning and DQNs}

Q-learning iteratively estimates the optimal Q-value function, $Q\left(s,a\right)=Q\left(s,a\right)+\alpha\left[r+\gamma\max_{a'}Q\left(s',a'\right)-Q\left(s,a\right)\right]$,
where $\alpha\in\left[0,1\right)$ is the learning rate and $\left[r+\gamma\max_{a'}Q\left(s',a'\right)-Q\left(s,a\right)\right]$
is the temporal-difference (TD) error. Convergence to $Q^{\pi^{*}}$
is guaranteed in the tabular (no approximation) case provided that
sufficient state/action space exploration is done; thus, tabulated
learning is not suitable for problems with large state spaces. Practical
TD methods use function approximators for the Q-value function such
as neural networks, i.e., deep Q-learning which exploits Deep Q-Networks
(DQNs) for Q-value approximation \cite{Mnih2015}.

RL can be unstable or even diverge when a nonlinear function approximator
such as a neural network is used to represent the Q-value function
\cite{NIPS1996_1269TDFunctionAproximator}. In order to overcome this
issue, DQNs rely on two key concepts, the \emph{experience replay
}and an iterative update that adjusts the Q-values towards \emph{target
values} that are only periodically updated.

The approximate Q-value function is parameterized using a deep neural
network, $Q\left(s,a;\phi\right)$, in which $\phi$ are the parameters
(weights) of the Q-network. To use experience replay, the agent's
experiences $\mu_{t}=\left(s_{t},a_{t},r_{t},s_{t+1}\right)$
are stored at each timestep $t$ in a data set $\mathcal{D}_{t}=\left\{ \mu_{1},\cdots,\mu_{t}\right\} $.
During learning, Q-learning updates are applied on samples (minibatches)
of experience $\left(s,a,r,s'\right)\sim U\left(\mathcal{D}\right)$,
drawn uniformly at random from the pool of stored samples. The Q-learning
update uses the following loss function:
\[
L\left(\phi\right)=\mathbb{E}_{\left(s,a,r,s'\right)\sim U\left(\mathcal{D}\right)}\left[\left(r+\gamma\max_{a'}Q\left(s',a';\phi^{-}\right)-Q\left(s,a;\phi\right)\right)^{2}\right],
\]
where $\phi^{-}$ are the network parameters used to compute the target.
The target network parameters $\phi^{-}$ are only updated with the
Q-network parameters $\phi$ every $C$ steps and remain fixed across
individual updates\footnote{Hereafter, for notation simplicity, $Q^{-}\left(s,a\right)$ and $Q\left(s,a\right)$
will be used instead of $Q\left(s,a;\phi^{-}\right)$ and $Q\left(s,a;\phi\right)$,
respectively.} \cite{Mnih2015}.

\subsection{Cooperative Perception Scenario}

In order to solve \eqref{First_optimization_problem}, the timeline
is splitted into two scales, a \emph{coarse scale} called time frames
and a \emph{fine scale} called time slots. At the beginning of each
time frame, the RSU associates vehicles into pairs and allocates RBs
to those pairs. The association and RB allocation stays fixed during
the whole frame which consists of $X$ time slots. At the beginning
of each time slot $t$, each vehicle selects the quadtree blocks to
be transmitted to its associated vehicle. By utilizing RL we can formulate
two different but interrelated RL problems: Vehicular RL and RSU RL.

\subsubsection{Vehicular RL}

In this RL problem, for a given association $nn'$ and RB allocation,
each vehicle $n$ acts as an RL-agent who wants to learn which quadtree
blocks to transmit to its associated vehicle $n'$ in order to maximize
the satisfaction of vehicle $n'$. Accordingly, the \emph{global state}
of the RL environment is defined as $\left\langle \mathcal{B}_{n}\left(t\right),\mathcal{I}_{n'}(t),v_{n},v_{n'},\boldsymbol{l}_{n}\left(t\right),\boldsymbol{l}_{n'}\left(t\right)\right\rangle $,
where $\mathcal{I}_{n'}\left(t\right)$ is the set of vehicle's $n'$
RoI weights, as per \eqref{eq:RoIWeights}, at time slot $t$. However,
this global state cannot be observed by vehicle $n$, where instead,
the\emph{ local observation} of vehicle $n$ is $\left\langle \mathcal{B}_{n}\left(t\right),v_{n},v_{n'},\boldsymbol{l}_{n}\left(t\right),\boldsymbol{l}_{n'}\left(t\right)\right\rangle $.
At every time slot $t$ and by utilizing this local observation, vehicle
$n$ takes an action $\boldsymbol{\sigma}_{n}\left(t\right)$, selecting
which quadtree blocks to be transmitted to its associated vehicle
$n'$, %
and accordingly receive a feedback (\emph{reward}) from vehicle $n'$
equal to $f_{n'n}\left(t\right)$. %
In a nutshell, the elements of the RL problem at each vehicle $n$
can be described as follows:
\begin{itemize}
\item Global~state: $\left\langle \mathcal{B}_{n}\left(t\right),\mathcal{I}_{n'}(t),v_{n},v_{n'},\boldsymbol{l}_{n}\left(t\right),\boldsymbol{l}_{n'}\left(t\right)\right\rangle $.
\item Local~observation: $\left\langle \mathcal{B}_{n}\left(t\right),v_{n},v_{n'},\boldsymbol{l}_{n}\left(t\right),\boldsymbol{l}_{n'}\left(t\right)\right\rangle $.
\item Action: $\boldsymbol{\sigma}_{n}\left(t\right)$.
\item Reward: $f_{n'n}\left(t\right)$.
\end{itemize}

\subsubsection{RSU RL}

The RSU acts as an RL-agent where the \emph{state} of this RL environment
is given by the location and velocity of all vehicles serviced by
the RSU, $\left\langle v_{n},\boldsymbol{l}_{n}\,\forall n\in\mathcal{N}\right\rangle $.
Based on this state at the beginning of each time frame, the RSU takes
the \emph{action} of vehicles association $E(t)$, and RB allocation
$\boldsymbol{\eta}(t)$. Then, once the time frame ends, each vehicle
will report back its mean satisfaction during the whole frame and
the RL \emph{reward }is computed as the mean of those feedbacks. In
a nutshell, the elements of the RL problem at the RSU can be summarized
as follows:
\begin{itemize}
\item State: $\left\langle v_{n},\boldsymbol{l}_{n}\,\forall n\in\mathcal{N}\right\rangle $.
\item Action: $E(t)$ and $\boldsymbol{\eta}(t)$.
\item Reward: $\frac{\sum_{n\in\mathcal{N}}\nicefrac{\left(\sum_{t=i}^{i+X}f_{n'n}\left(t\right)\right)}{X}}{|\mathcal{N}|}$.
\end{itemize}
In order to solve these two RL problems, the DQN algorithm \cite{Mnih2015}
can be used. However, despite its success in domains with high-dimensional
state space such as our domain, its application to high dimensional,
discrete action spaces is still arduous, because within DQN, the Q-value
for each possible action should be estimated before deciding which
action to take. Furthermore, the number of actions that need to be
explicitly represented grows exponentially with increasing action
dimensionality \cite{ActionBranching}.

At this point, we note that our two RL problems suffer from the high
dimensionality of action spaces. Specifically, within the RSU RL problem,
the RSU needs to select $E(t)$ and $\boldsymbol{\eta}(t)$: The association
matrix $E(t)$ is of size $N\times N$, and due to our one-to-one
association assumption, the number of possible actions for the association
problem would be $\Pi_{n=1}^{\left\lfloor \nicefrac{N}{2}\right\rfloor }\left(2n-1\right)$.
Moreover, the RB allocation matrix $\boldsymbol{\eta}(t)$ is of size
$N\times K$, as a result, the number of possible actions is $K^{N}$,
assuming that each vehicle is allocated only 1 RB. Similarly, within
the vehicular RL problem, each vehicle needs to select $\boldsymbol{\sigma}_{n}\left(t\right)$
whose dimension is $|\mathcal{B}_{n}|_{\text{max}}\times1$, yielding
a total number of possible actions equal to $2^{|\mathcal{B}_{n}|_{\text{max}}}$.

This large number of actions can seriously affect the learning behavior
of the available discrete-action reinforcement learning algorithms
such as DQN, because large action spaces are difficult to explore
efficiently and thus successful training of the neural networks becomes
intractable \cite{Lillicrap2015}.

\section{Overcoming the Large Action Space Problem\label{sec:Overcoming-the-huge}}

\begin{figure}[t]
\centering \includegraphics[width=0.9\textwidth]{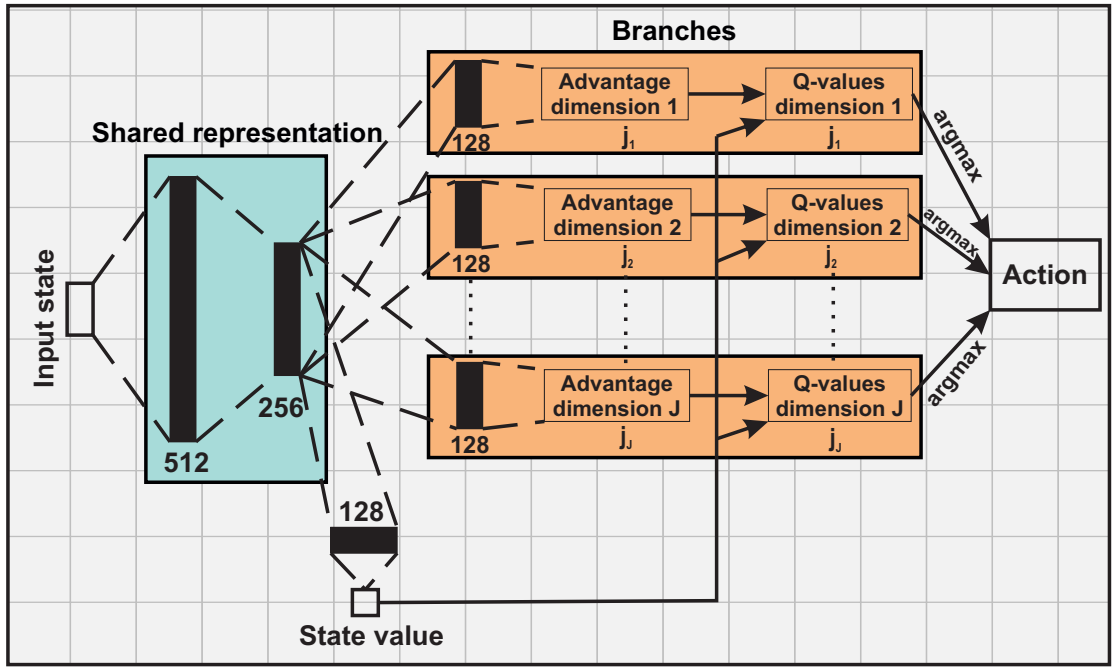}
\caption{The BDQ neural network architecture utilized for both RSU and vehicular
RL agents.}
\label{fig:NN}
\end{figure}

Recently, the authors in \cite{ActionBranching} have introduced a
new agent called branching dueling Q-network (BDQ). The resulting
neural network architecture allows to distribute the representation
of the action dimensions across individual network branches while
maintaining a shared module that encodes a latent representation of
the input state and helps to coordinate the branches. This architecture
is represented in Fig.\,\ref{fig:NN}. Remarkably, this neural network
architecture exhibits a linear growth of the network outputs with
increasing action space as opposed to the combinatorial growth experienced
in traditional DQN network architectures.

Here, we adopt these BDQ agents from \cite{ActionBranching} within
our RL problems. As a result, the neural network at the RSU agent
will have $N$ branches\footnote{$N-1$ branches if $N$ is odd.} constructed
as follows:
\begin{itemize}
\item $\left\lfloor \nicefrac{N}{2}\right\rfloor $ branches corresponding
to the association action with each branch having $j_{i}=N-2i+1$
sub-actions, where $i$ is the branch ID. For example, let us consider
a simplified scenario with $N=6$, then $\left\lfloor \nicefrac{N}{2}\right\rfloor =3$
vehicular pairs could be formed: the first branch representing the
first vehicle would have $N-2\cdot(1)+1=5$ candidate vehicles to
pair with, while for the second branch the candidates are reduced
to 3 and so on. This leads to a unique vehicular association for any
combination of sub-actions selected at each of the branches. For instance,
an action of $(1,1,1)$ implies that $E=\left[\begin{array}{cccccc}
0 & 1 & 0 & 0 & 0 & 0\\
1 & 0 & 0 & 0 & 0 & 0\\
0 & 0 & 0 & 1 & 0 & 0\\
0 & 0 & 1 & 0 & 0 & 0\\
0 & 0 & 0 & 0 & 0 & 1\\
0 & 0 & 0 & 0 & 1 & 0
\end{array}\right],$ and an action of $(3,2,1)$ would mean that $E=\left[\begin{array}{cccccc}
0 & 0 & 0 & 1 & 0 & 0\\
0 & 0 & 0 & 0 & 1 & 0\\
0 & 0 & 0 & 0 & 0 & 1\\
1 & 0 & 0 & 0 & 0 & 0\\
0 & 1 & 0 & 0 & 0 & 0\\
0 & 0 & 1 & 0 & 0 & 0
\end{array}\right]$.
\item $\left\lfloor \nicefrac{N}{2}\right\rfloor $ branches corresponding
to the RB allocation with each branch having ${K \choose 2}$ sub-actions,
knowing that each associated pair is allocated 2 orthogonal RBs (one
for each vehicle).
\end{itemize}
The aftermath of using the BDQ agent is that, in order to select an
association action $E(t)$, the Q-value needs to be estimated for
$\sum_{n=1}^{\left\lfloor \nicefrac{N}{2}\right\rfloor }\left(2n-1\right)$
actions instead of for $\Pi_{n=1}^{\left\lfloor \nicefrac{N}{2}\right\rfloor }\left(2n-1\right)$
with a non-branching network architecture. Similarly, selecting an
RB allocation $\boldsymbol{\eta}(t)$, requires the Q-value estimation
of $\frac{N}{2}\times{K \choose 2}$ actions instead of the ${K \choose 2}^{\nicefrac{N}{2}}$values
involved in a traditional DQN architecture. Equivalently, by utilizing
the BDQ agent within our vehicular RL problem, for the message content
selection $\boldsymbol{\sigma}_{n}\left(t\right)$, the Q-value needs
to be estimated for $2\times|\mathcal{B}_{n}|_{\text{max}}$ actions
only instead of for $2^{|\mathcal{B}_{n}|_{\text{max}}}$ actions.

\subsection{Training a BDQ Agent within The Cooperative Perception Scenario}

For training the RSU and vehicular agents, DQN is selected as the
algorithmic basis. Thus, at the beginning of each RSU episode, a random
starting point of an arbitrary trajectory of vehicles is selected,
resulting in a an indiscriminate state $\left\langle v_{n},\boldsymbol{l}_{n}\,\forall n\in\mathcal{N}\right\rangle $
observed by the RSU. Here, this state is the input to the BDQ agent
(neural network) available at the RSU. Then, with probability $\epsilon$,
this BDQ agent randomly selects the association $E\left(t\right)$
and RB allocation $\boldsymbol{\eta}(t)$ actions, and with probability
$1-\epsilon$, it will select the action having the maximum Q-value\footnote{The value of $\epsilon$ is reduced as the learning proceeds till
it reaches $0$, in order to ensure an efficient exploration-exploitation
balance.} (as determined by the output of the neural network).

For any action dimension $i\in\left\{ 1,\dots,J\right\} $ with $\left|\mathcal{A}_{i}\right|=j_{i}$
discrete sub-actions, the Q-value of each individual branch at state
$s\in\mathcal{S}$ and sub-action $a_{i}\in\mathcal{A}_{i}$ is expressed
in terms of the common state value $V\left(s\right)$ and the corresponding
state-dependent sub-action advantage $A_{i}\left(s,a_{i}\right)$
by \cite{ActionBranching}:

\begin{equation}
Q_{i}\left(s,a_{i}\right)=V\left(s\right)+\left(A_{i}\left(s,a_{i}\right)-\frac{1}{j_{i}}\sum_{a'_{i}\in\mathcal{A}_{i}}A_{i}\left(s,a'_{i}\right)\right).
\end{equation}

After the action is determined, the RSU forwards the association and
RB allocation decision to the corresponding vehicles. This association
and RB allocation decision will hold for the upcoming $X$ time slots.
Once the RSU decision has been conveyed to the vehicles, each vehicle
$n$ can compute its local observation $\left\langle \mathcal{B}_{n}\left(t\right),v_{n},v_{n'},\boldsymbol{l}_{n}\left(t\right),\boldsymbol{l}_{n'}\left(t\right)\right\rangle $.
Note here that, this local observation constitutes the input for the
BDQ agent running at vehicle $n$. Furthermore, an $\epsilon-$greedy
policy is also employed at each vehicle, thus random sensory blocks
will be selected for transmission with probability $\epsilon$, and
the sensory blocks which maximizes the Q-value with probability $1-\epsilon$.
Then, the resulting sensory blocks will be scheduled for transmitted
over the allocated RB to the associated vehicle. Notice that, the
associated vehicle might only receive a random subset of these blocks
depending on the data rate $R_{nn'}\left(t\right)$ as per \eqref{eq:rate}.
It will then calculate its own satisfaction $f_{n'n}\left(t\right)$
with the received blocks according to \eqref{eq:VehicularSatisfaction}
and feed this value back as a reward to vehicle $n$. Vehicle $n$
receives the reward, observes the next local observation and stores
this experience $\mu_{t}^{n}=\left(s_{t},a_{t},r_{t},s_{t+1}\right)$
in a data set $\mathcal{D}_{t}^{n}=\left\{ \mu_{1}^{n},\cdots,\mu_{t}^{n}\right\} $.
After $X$ time slots, each vehicle will feedback its average received
reward during the whole frame to the RSU that will calculate the mean
of all the received feedbacks and use the result as its own reward
for the association and RB allocation action. The RSU stores its own
experience, $\mu_{m}^{\text{RSU}}=\left(s_{m},a_{m},r_{m},s_{m+1}\right)$,
in a data set $\mathcal{D}_{m}^{\text{RSU}}=\left\{ \mu_{1}^{\text{RSU}},\cdots,\mu_{m}^{\text{RSU}}\right\} $,
where $m$ is the frame index. A new RSU episode begins every $Z$
frames.

Once an agent has collected a sufficient amount of experience, the
training process of its own neural network starts. First, samples
of experience (mini-batch) are drawn uniformly at random from the
pool of stored samples, $\left(s,a,r,s'\right)\sim U\left(\mathcal{D}\right)$
\footnote{Super/subscript is omitted here for notation simplicity.}.
Using these samples, the loss function within the branched neural
network architecture of the BDQ agent is calculated as follows \cite{ActionBranching}:

\begin{equation}
L\left(\phi\right)=\mathbb{E}_{\left(s,a,r,s'\right)\sim U\left(\mathcal{D}\right)}\left[\frac{1}{J}\sum_{i}\left(y_{i}-Q_{i}\left(s,a_{i}\right)\right)^{2}\right],\label{eq:LossFunction}
\end{equation}
where $i$ is the branch ID, $J$ is the total number of branches,
and $a$ denotes the joint-action tuple $\left(a_{1},\cdots,a_{i},\cdots,a_{J}\right)$.
Moreover, $y_{i}=r+\gamma\frac{1}{J}\sum_{i}Q_{i}^{-}\left(s',\text{arg max}_{a'_{i}\in\mathcal{A}_{i}}Q_{i}\left(s',a'_{i}\right)\right)$
in \eqref{eq:LossFunction} represents the temporal difference targets\footnote{For a complete discussion on the choice of the loss function and its
components, please refer to \cite{ActionBranching}.}. Finally, a gradient descent step is performed on $L\left(\phi\right)$
with respect to the network parameters $\phi$. The training process
of the BDQ agents is summarized in Algorithm\,\ref{alg:Training BDQ}.
\begin{algorithm}[t]
\begin{algorithmic}[1]\footnotesize

\STATE  \textbf{Initialize }the replay memory of each agent to a
fixed buffer size.

\STATE  \textbf{Initialize }each agent's neural network with random
weights $\phi$.

\STATE  \textbf{Initialize }each agent's target neural network with
weights $\phi^{-}=\phi$.

\STATE    \textbf{foreach} RSU episode \textbf{do}

 \begin{ALC@g} \STATE  Reset the RSU environment by selecting random
trajectories for all vehicles within the junction scenario.

\STATE  The RSU observes its current state $\left\langle v_{n},\boldsymbol{l}_{n}\,\forall n\in\mathcal{N}\right\rangle $.

\STATE \textbf{foreach }$Z$ frames \textbf{do}

 \begin{ALC@g} \STATE  With probability $\epsilon$, the RSU agent
selects a random association and RB allocation action, otherwise the
action with maximum Q-value is selected.

\STATE  This action (decision) is forwarded to the corresponding
vehicles.

\STATE \textbf{foreach }$X$ slots at each vehicle\textbf{ do}

 \begin{ALC@g} \STATE  Vehicle $n$ computes its local observation
$\left\langle \mathcal{B}_{n}\left(t\right),v_{n},v_{n'},\boldsymbol{l}_{n}\left(t\right),\boldsymbol{l}_{n'}\left(t\right)\right\rangle $.

\STATE  With probability $\epsilon$, it selects random sensory blocks
to be transmitted to its associated vehicle, otherwise the sensory
blocks with maximum Q-value are selected.

\STATE  Transmit over the allocated RB to the associated vehicle;
As per rate $R_{nn'}\left(t\right)$ in \eqref{eq:rate} only a random
subset of these blocks will be received.

\STATE  It calculates its own satisfaction $f_{nn'}\left(t\right)$
as per \eqref{eq:VehicularSatisfaction} and feeds it back as a reward
to the associated vehicle.

\STATE  Receive the reward, observe the next local observation and
store this experience $\left(s_{t},a_{t},r_{t},s_{t+1}\right)$ in
its replay memory.

\STATE \textbf{if }vehicle $n$ has collected a sufficient amount
of experiences \textbf{do}

 \begin{ALC@g} \STATE  Vehicle $n$ samples uniformly a random mini-batch
of experiences $\mu^{n}$ from its replay memory.

\STATE  It performs a gradient decent step on $L\left(\phi\right)$
w.r.t. $\phi$, using the samples.

 \end{ALC@g}\STATE  \textbf{end} \textbf{if}

 \end{ALC@g}\STATE  \textbf{end} \textbf{for}

\STATE  Each vehicle feeds back its average received reward during
the whole frame to the RSU.

\STATE  The RSU calculates the mean of all the received feedbacks
and use the result as its own reward.

\STATE  The RSU stores its own experience, $\left(s_{i},a_{i},r_{i},s_{i+1}\right)$,
in its replay memory.

\STATE \textbf{if }the RSU collected a sufficient amount of experiences
\textbf{do}

 \begin{ALC@g} \STATE  Sample uniformly a random mini-batch of experiences
from its replay memory.

\STATE  Using these samples, a gradient decent step is performed
on $L\left(\phi\right)$ w.r.t. $\phi$.

 \end{ALC@g}\STATE  \textbf{end} \textbf{if}

 \end{ALC@g}\STATE  \textbf{end for}

 \end{ALC@g}\STATE  \textbf{end for}

\end{algorithmic}

\caption{\label{alg:Training BDQ} Training a BDQ agent for cooperative perception}
\end{algorithm}

\section{Federated RL\label{sec:Federated-RL}}

We now observe that, so far, each vehicle $n$ has only leveraged
its own experience to train its BDQ agent independently. Therefore,
in order to have a resilient agent that performs well in different
situations, the training process should run for a sufficient amount
of time for the vehicle to gain a broad experience. Alternatively,
vehicles could periodically share their trained models with each other
to enhance the training process and obtain a better model in a shorter
amount of time.

For that purpose, we investigate the role of federated RL \cite{FRL}
where different agents (vehicles) collaboratively train a global model
under the orchestration of a central entity (RSU), while keeping the
training data (experiences) decentralized \cite{FLmcmahan17a,FLSumudu}.
Instead of applying federated learning (FL) within a supervised learning
task, in this work, we investigate the use of FL for reinforcement
learning within our cooperative perception vehicular RL problem. In
particular, at the end of every time frame $m$, each vehicle $n$,
under the service of the RSU, updates (trains) its local model (neural
network weights) $\phi_{m}^{n}$ based on its local experiences, by
performing a gradient descent step on $L\left(\phi_{m}^{n}\right)$
as per \eqref{eq:LossFunction}. Next, each vehicle shares this updated
model with the RSU which computes a global model by aggregating all
the received models as follows:
\[
\phi_{m}^{*}=\frac{1}{N}\sum_{n}\phi_{m}^{n},
\]
where $\phi_{m}^{*}$ is the global model computed by the RSU at time
frame $m$. After computing the global model, the RSU broadcasts $\phi_{m}^{*}$
back to the vehicles under its service, where each vehicle replaces
its local model with $\phi_{m}^{*}$. Algorithm\,\ref{alg:FRLAlgo}
summarizes the entire FRL process within our cooperative perception
scenario.
\begin{algorithm}[t]
\begin{algorithmic}[1]\footnotesize

\STATE   \textbf{foreach} frame $m$ \textbf{do}

\begin{ALC@g} \STATE   \textbf{At each vehicle $n$ served by the
RSU}

 \begin{ALC@g} \STATE  Perform a gradient descent step on $L\left(\phi_{m}^{n}\right)$
as per \eqref{eq:LossFunction}.

\STATE  Update the local model $\phi_{m}^{n}$.

\STATE  Share $\phi_{m}^{n}$ with the RSU.

 \end{ALC@g}\STATE \textbf{At the RSU}

 \begin{ALC@g} \STATE  Aggregate the received models according to
$\phi_{m}^{*}=\frac{1}{N}\sum_{n}\phi_{m}^{n}$.

\STATE  Broadcast $\phi_{m}^{*}$ back to the vehicles.

 \end{ALC@g} \end{ALC@g} \STATE \textbf{end} \textbf{for}

\end{algorithmic}

\caption{\label{alg:FRLAlgo} FRL for vehicular cooperative perception}
\end{algorithm}

\section{Simulation Results and Analysis\label{sec:Numerical-results}}

Simulations are conducted based on practical traffic data to demonstrate the effectiveness of the proposed approach. 
A traffic light regulated junction scenario is considered.
Several junction scenes of random vehicles' mobility traces were generated using the Simulation of Urban MObility (SUMO) framework \cite{SUMO}.
Each scene spans a $30000$ time slots and consists of a total of 30 vehicles entering and exiting the coverage area of a single RSU. The vehicles are of different dimensions to mimic assorted cars, buses, and trucks.
Unless stated otherwise, the simulation parameters are listed in Table~\ref{tab:sim_par}\footnote{Simulations show the benefits of our method even
for vehicles with error-free sensors.}.
\begin{table}[t]
\caption{Simulation parameters. \label{tab:sim_par}}
\centering{}%
\begin{tabular}{|c|c||c|c|}
\hline 
\textbf{Parameter} & \textbf{Value} & \textbf{Parameter} & \textbf{Value}\tabularnewline
\hline 
\hline 
$K$ & $10$ & $N_{0}$ & $-174$ dBm/Hz\tabularnewline
\hline 
$\omega$ & $180$ kHz & $P$ & $10$ dBm\tabularnewline
\hline 
$\tau$ & $2$ ms & $t_{\text{int}}$ & $2$ sec\tabularnewline
\hline 
$M$ & $100$ bytes & $L$ & 5\tabularnewline
\hline 
$\lambda_{n}$ & 1 & $r$ & 20\tabularnewline
\hline 
$X$ & $5$ slots & $Z$ & $10$ frames\tabularnewline
\hline 
\end{tabular}
\end{table}

Moreover, the hyperparameters used for training the RSU and vehicular
agents are discussed next. Common to all agents, training always starts
after the first $1000$ simulation steps; subsequently, for each simulation
time step a training step will be run. Adam optimizer is used with
a learning rate of $10^{-4}$. Training is performed with a minibatch
size of $64$ and a discount factor $\gamma=0.99$. In addition, the
target network is updated every $1000$ time steps. A rectified non-linearity
(ReLU) is used for all hidden layers and a linear activation is used
on the output layers, for all neural networks. Each neural network
is comprised of two hidden layers with $512$ and $256$ units in
the shared network module and of one hidden layer per branch with
$128$ units. Finally, a buffer size of $10^{6}$ is set for the replay
memory of each agent.

\begin{figure}[t]
\centering \includegraphics[width=0.9\textwidth]{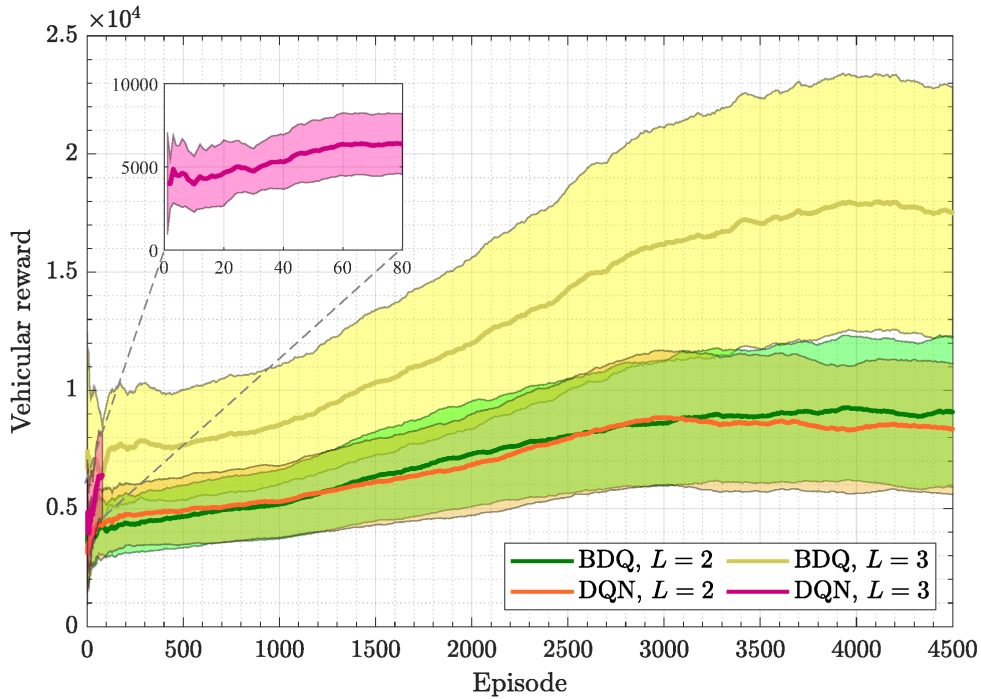}
\caption{Learning curves for the vehicular RL environment. The solid lines
represent the average over all the vehicles, where the learning curve
of each vehicle is smoothed by the moving average over a window size
of 1000 episodes, while the shaded areas show the 90\% confidence
interval over the vehicles.}
\label{fig:Baslines}
\end{figure}

First of all, we verify whether the BDQ agent is able to deal with
the huge action space problem without experiencing any notable performance
degradation when compared to a classical DQN agent. For this purpose,
we alter the size of the action space of the vehicular RL problem
by increasing the maximum quadtree resolution $L$. Note that, when
$L=2,$ the maximum number of blocks available is $\frac{1-4^{L}}{1-4}=5$,
resulting in a total number of actions of $2^{5}=32$, whereas when
$L=3$, the maximum number of blocks available is $21$, leading to
a total number of $2^{21}\thickapprox2\times10^{6}$ actions, assuming
that each vehicle $n$ only transmits blocks within its $\mathcal{B}_{n}^{\text{c}}$.
Fig.\,\ref{fig:Baslines} shows the learning curve of both BDQ and
DQN agents, for each case of $L$. When $L=2$ (small action space),
the learning curves of both BDQ and DQN agents are comparable and
they learn with the same rate. However, when $L$ increases to $3$
(large action space), the training process of the DQN agent could
not be completed because it was computationally expensive. This is
due to the large number of actions that need to be explicitly represented
by the DQN network and hence, the extreme number of network parameters
that must be trained at every iteration. The BDQ agent, however, performs
well and shows robustness against huge action spaces, which demonstrates
its suitability to overcome the scalability problems faced by other
forms of RL.

\begin{figure}[t]
\centering \includegraphics[width=0.9\textwidth]{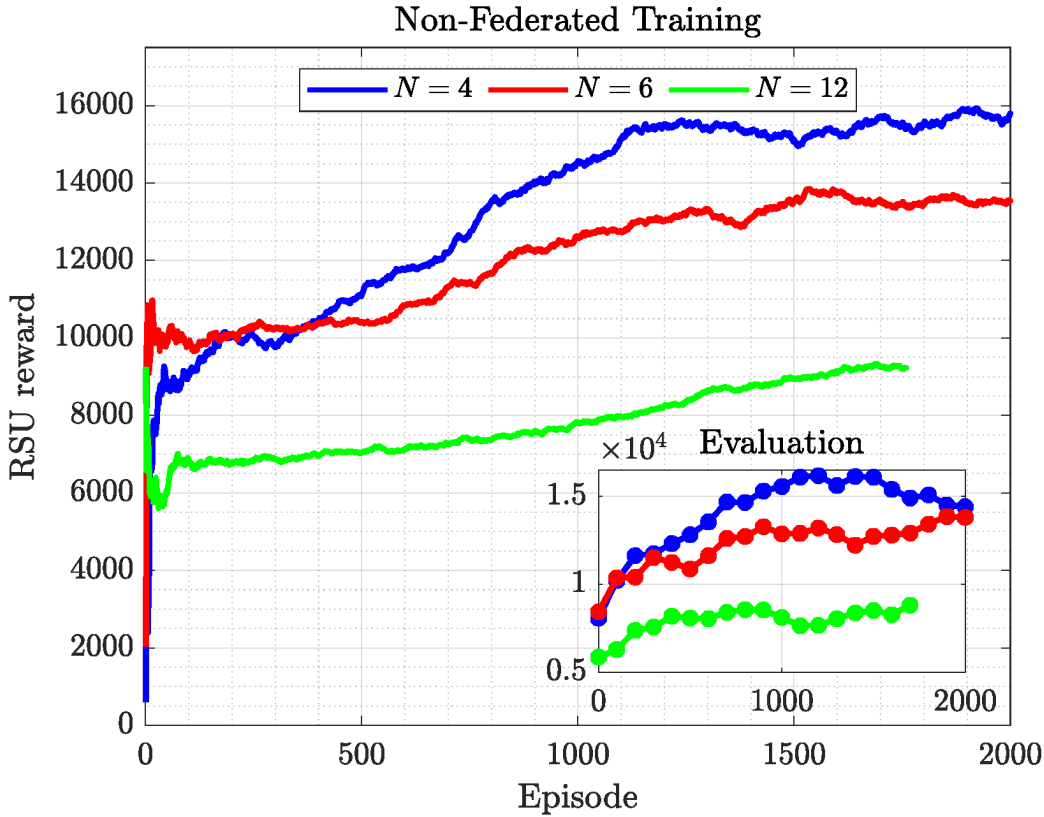}
\caption{Training and evaluation curves of the non-Federated scenario for the
RSU agent for different $N$. Each line is smoothed by the moving
average over a window size of 500 episodes.}
\label{fig:NonFederatedRSUTraining}
\end{figure}

Next, in Fig.\,\ref{fig:NonFederatedRSUTraining}, we study the training
progress of the RSU agent within the non-federated scenario for different
values of $N$, where $N$ is the maximum number of vehicles that
could be served by the RSU. Fig.\,\ref{fig:NonFederatedRSUTraining}
demonstrates how the RSU reward increases gradually with the number
of training episodes, i.e., the RSU and vehicles learn a better association,
RB allocation and message content selection over the training period.
However, it can be noted that the rate of increase of the RSU reward
decreases as the number of served vehicles $N$ increases and, hence,
more episodes are required to reach the same performance. The latter
is motivated by the inflation in the state space of the RSU agent,
which would require more episodes to be explored. Moreover, evaluations
were conducted every 100 episodes of training for 10 episodes with
a greedy policy. Fig.\,\ref{fig:NonFederatedRSUTraining} shows the
progress of the evaluation process during training and verifies that
agents learn better policies along the training duration.

\begin{figure}[t]
\centering %
\begin{tabular}{cc}
\includegraphics[width=0.45\textwidth]{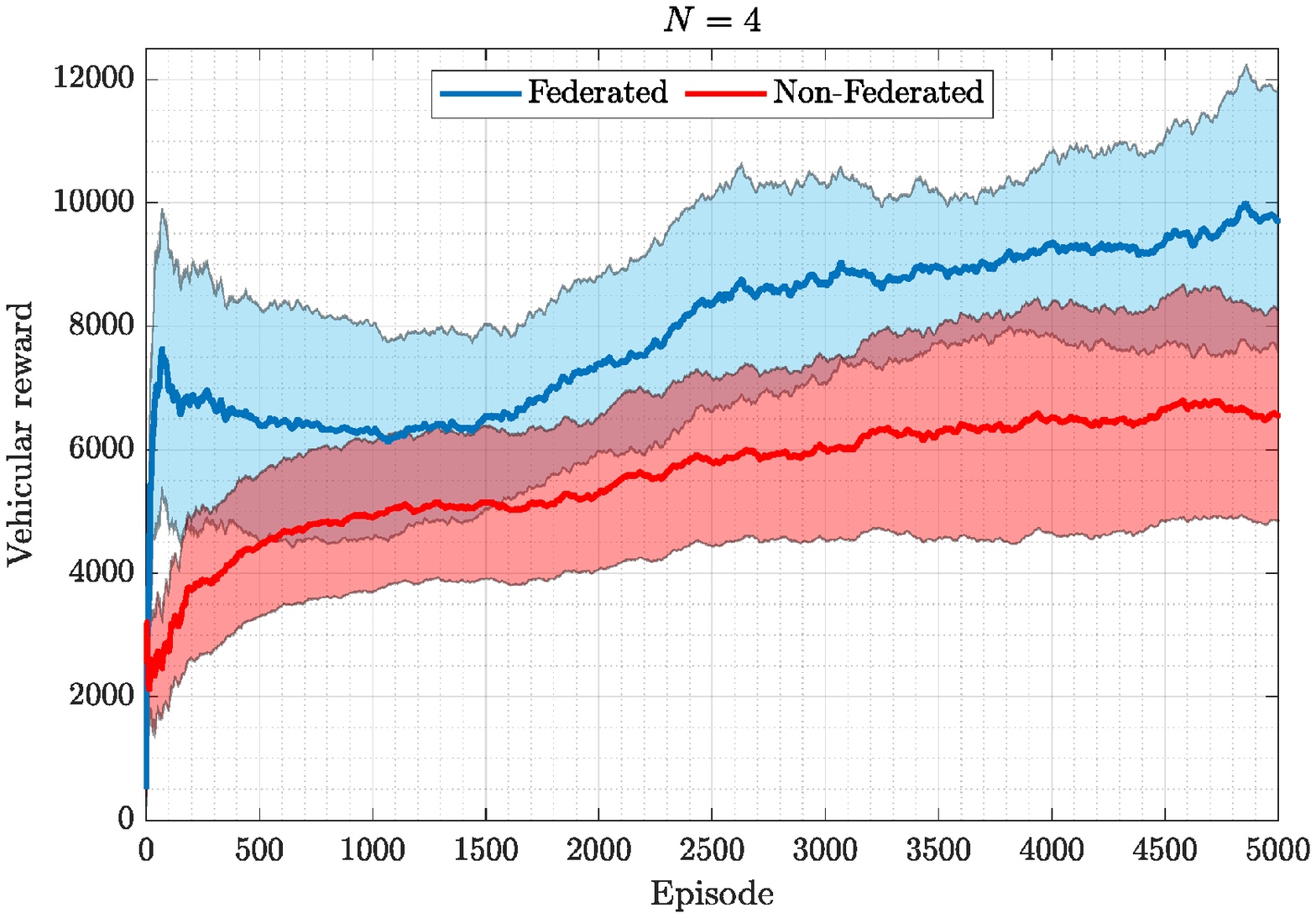} & \includegraphics[width=0.45\textwidth]{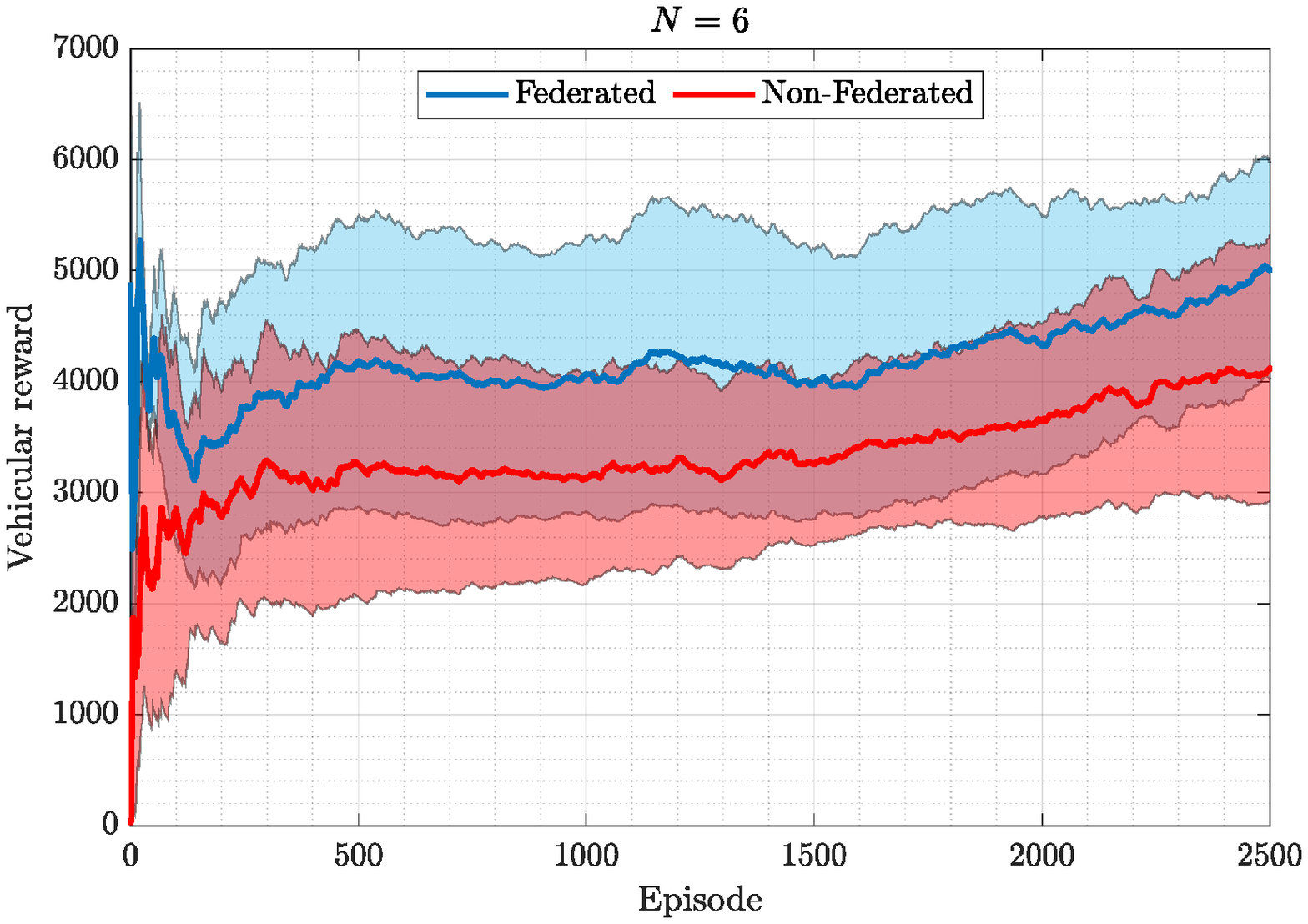}\tabularnewline
(a) & (b)\tabularnewline
\multicolumn{2}{c}{\includegraphics[width=0.45\textwidth]{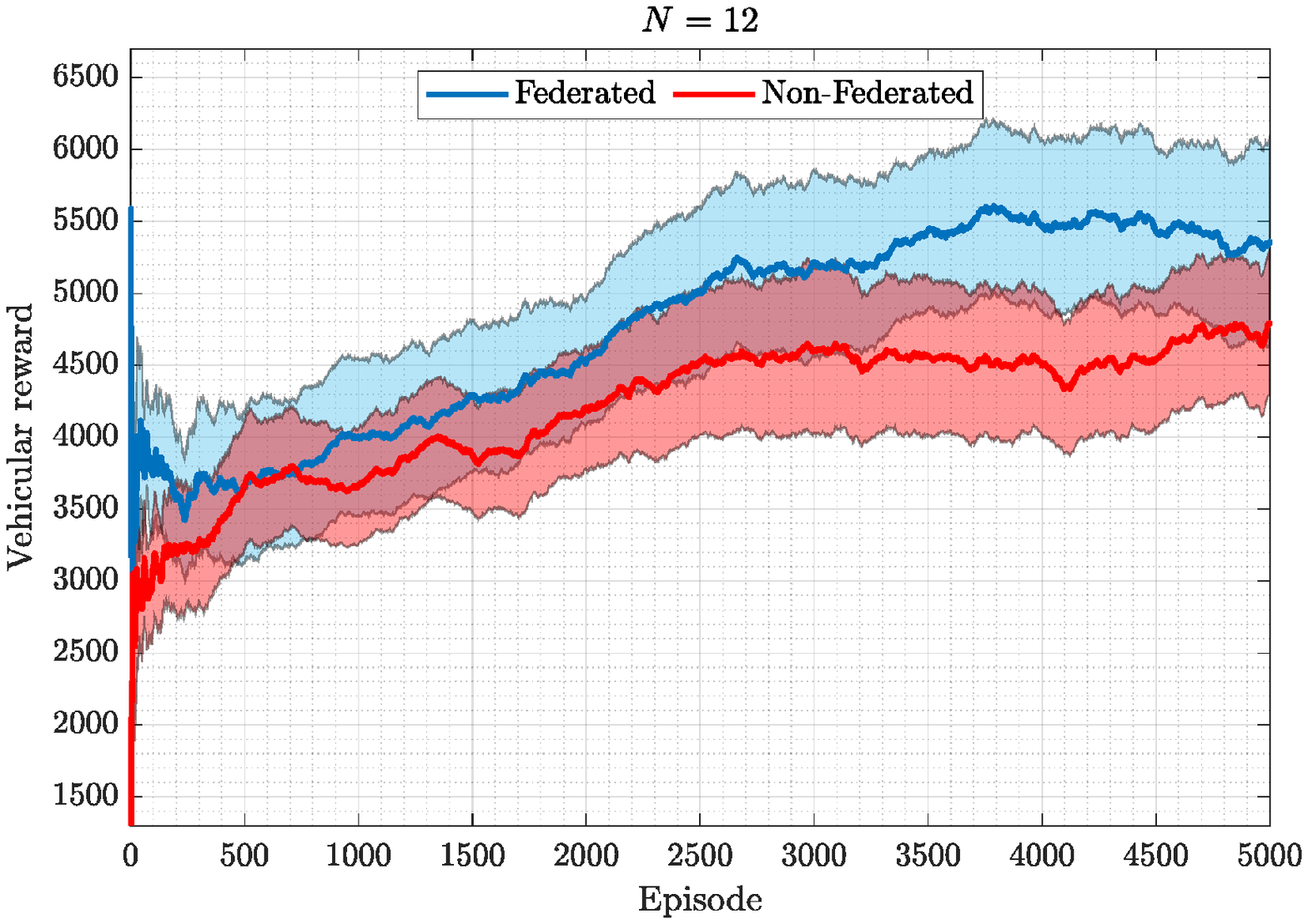}}\tabularnewline
\multicolumn{2}{c}{(c)}\tabularnewline
\end{tabular}\caption{Learning curves for the federated vs non-federated scenarios of vehicular
cooperative perception environment with $L=5$. The solid lines represent
the average over all the vehicles, where the learning curve of each
vehicle is smoothed by the moving average over a window size of 1000
episodes, while the shaded areas show the 90\% confidence interval
over the vehicles.}
\label{fig:FedVsNon}
\end{figure}

In Fig.\,\ref{fig:FedVsNon}, we compare the evolution of the training
process both for the federated and non-federated scenarios, and for
different values of $N$. From this figure, we observe that for the
same training period, if compared to the non-federated scenario, the
federated scenario achieves better rewards, and, hence, better policies
over all vehicles. This result corroborates that FL algorithms are
instrumental in enhancing and boosting the RL training process.

\begin{figure}[t]
\centering \includegraphics[width=0.9\textwidth]{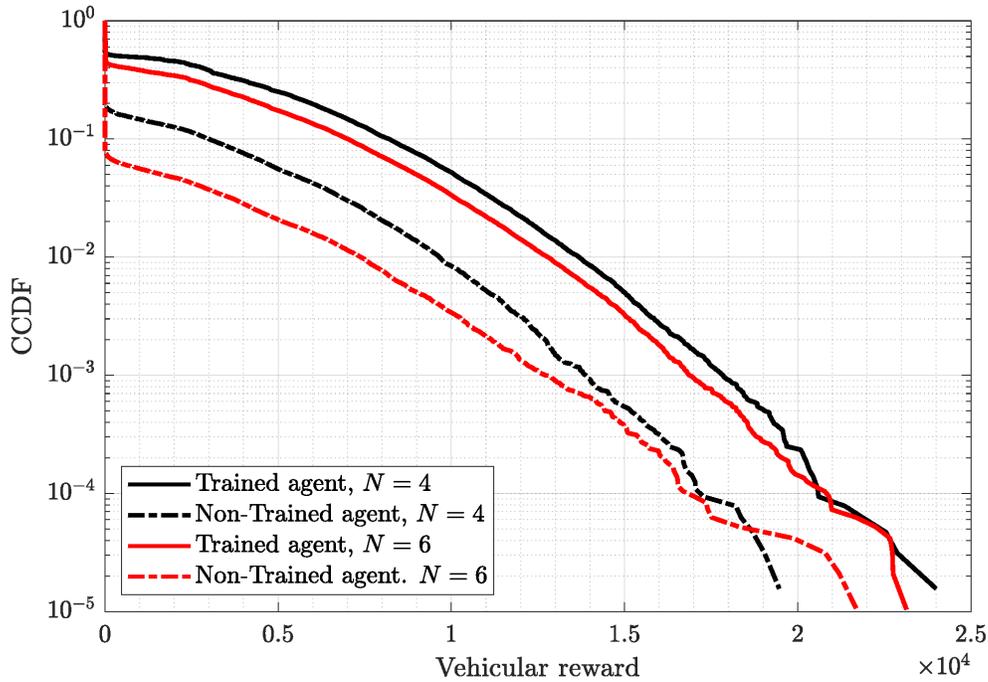}
\caption{The CCDF of the vehicular reward achieved by trained and non-trained
agents for different $N$.}
\label{fig:TrainedvsNonTrainedAgent}
\end{figure}

\begin{figure}[t]
\centering %
\begin{tabular}{cc}
\includegraphics[width=0.45\textwidth]{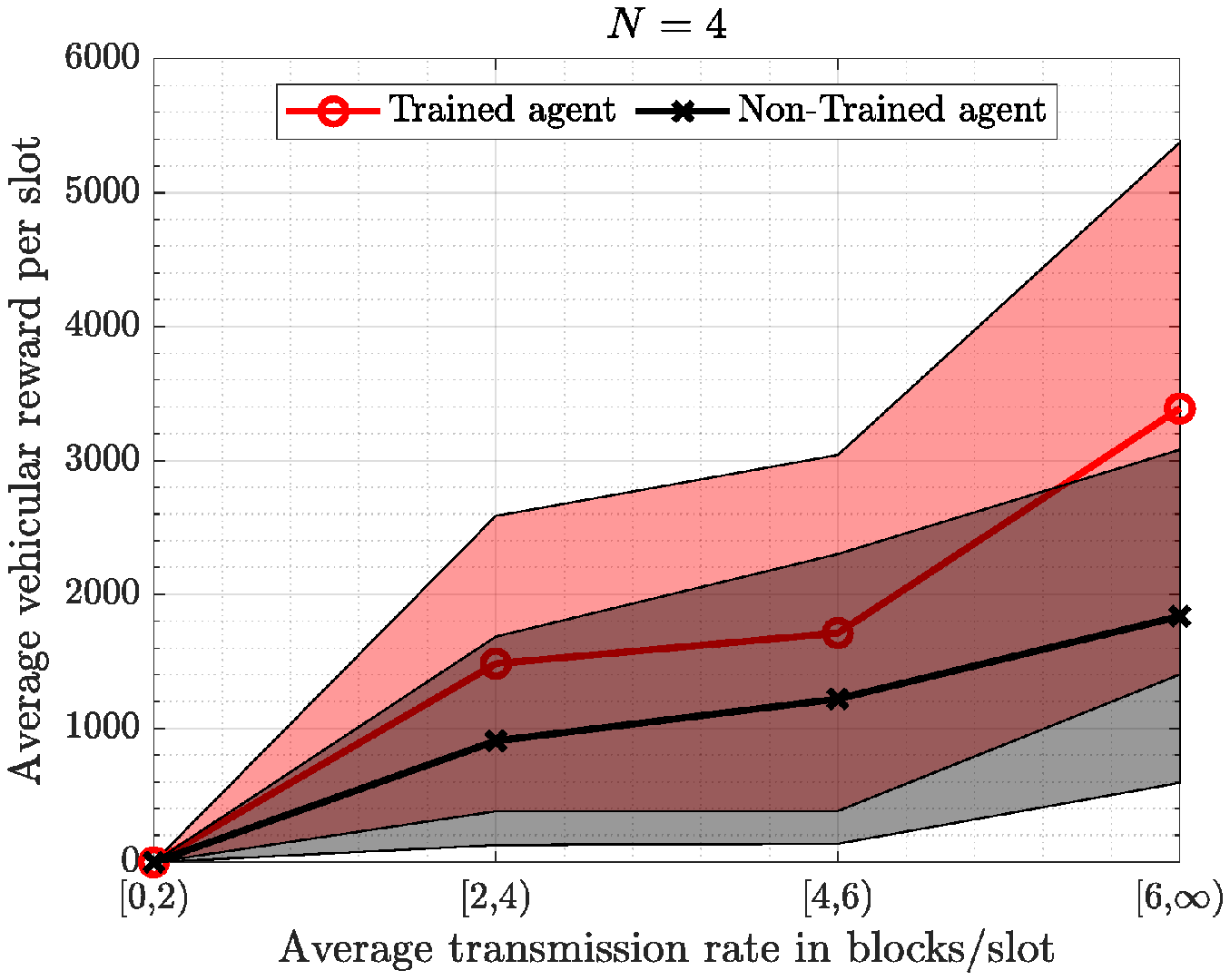} & \includegraphics[width=0.45\textwidth]{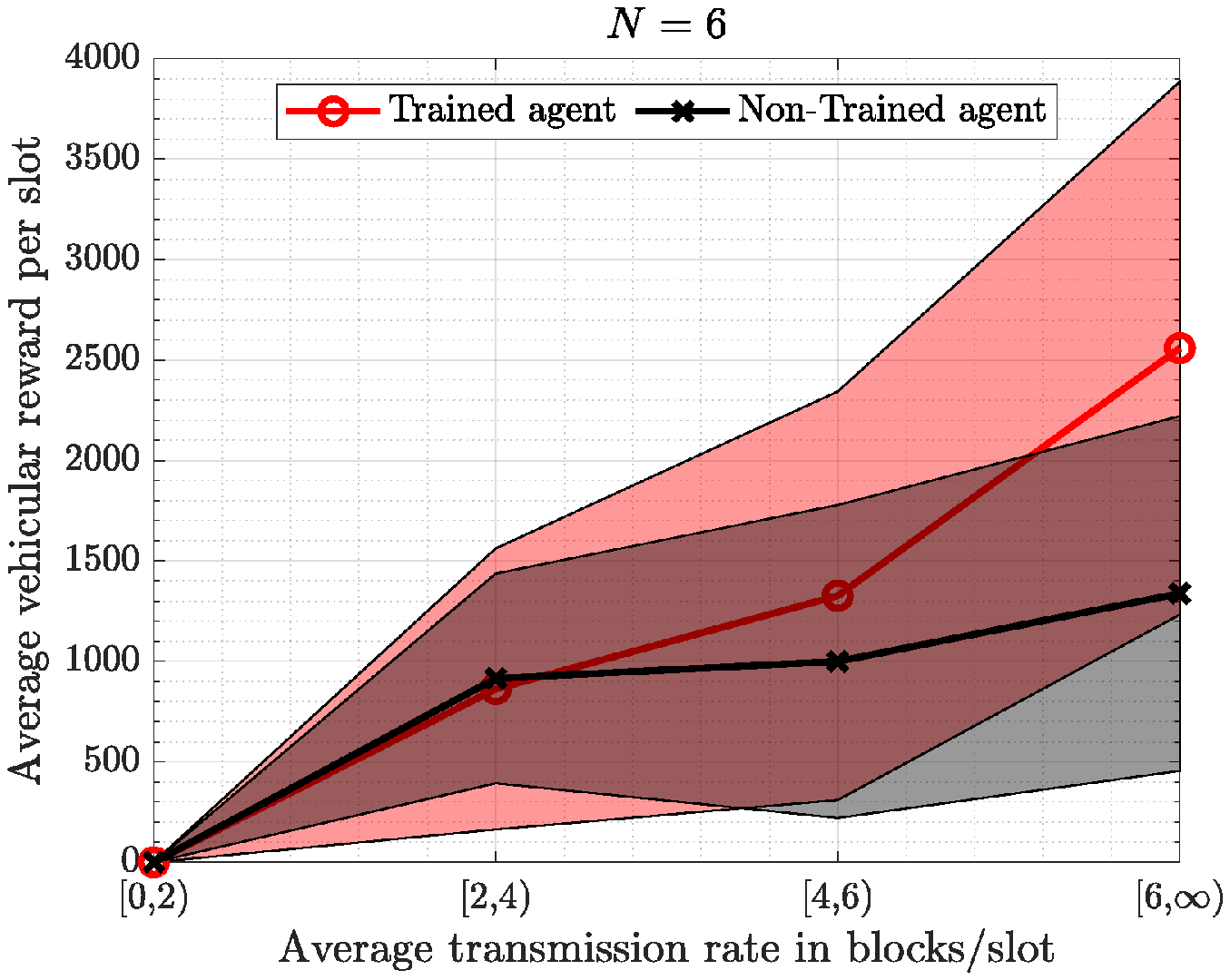}\tabularnewline
(a) & (b)\tabularnewline
\end{tabular}\caption{The average transmission rate vs the average vehicular reward achieved
by trained and non-trained agents for different $N$. The solid lines
represents the mean of the vehicular reward within each range of the
transmission rate, while the shaded areas show its standard deviation.}
\label{fig:RateVsReward}
\end{figure}

Once the trained RSU and vehicular agents have been obtained, those
agents are deployed within a newly generated vehicular mobility trajectory
scenario that runs for $20000$ slots. Fig.\,\ref{fig:TrainedvsNonTrainedAgent}
shows the complementary cumulative distribution function (CCDF) of
the vehicular rewards of all the vehicles and different $N$ values
under two scenarios: using trained vs. non-trained agents that select
their actions randomly. We can see by simple inspection, that the
vehicular reward distribution achieved by trained agents is superior
to the non-trained cases. This result holds both for $N=4$ and $N=6$.
Moreover, Fig.\,\ref{fig:RateVsReward} shows the average achieved
vehicular reward versus the average transmission rate. Note that,
for a given range of transmission rates, a trained agent achieves
a better vehicular reward than a non-trained agent both for $N=4$
and $N=6$, e.g., trained agent can achieve on average about $60\%$
and $40\%$ more reward for a given range of transmission rates when
$N=4$ and $N=6$ respectively. Also, the trained agent can achieve
the same vehicular reward with a lower transmission rate compared
to the non-trained agent. In summary, leveraging RL, the RSU and vehicular
agents learned how to take better actions for association, RB allocation
and message content selection, so as to maximize the achieved vehicular
satisfaction with the received sensory information.

\begin{figure}[t]
\centering \includegraphics[width=0.9\textwidth]{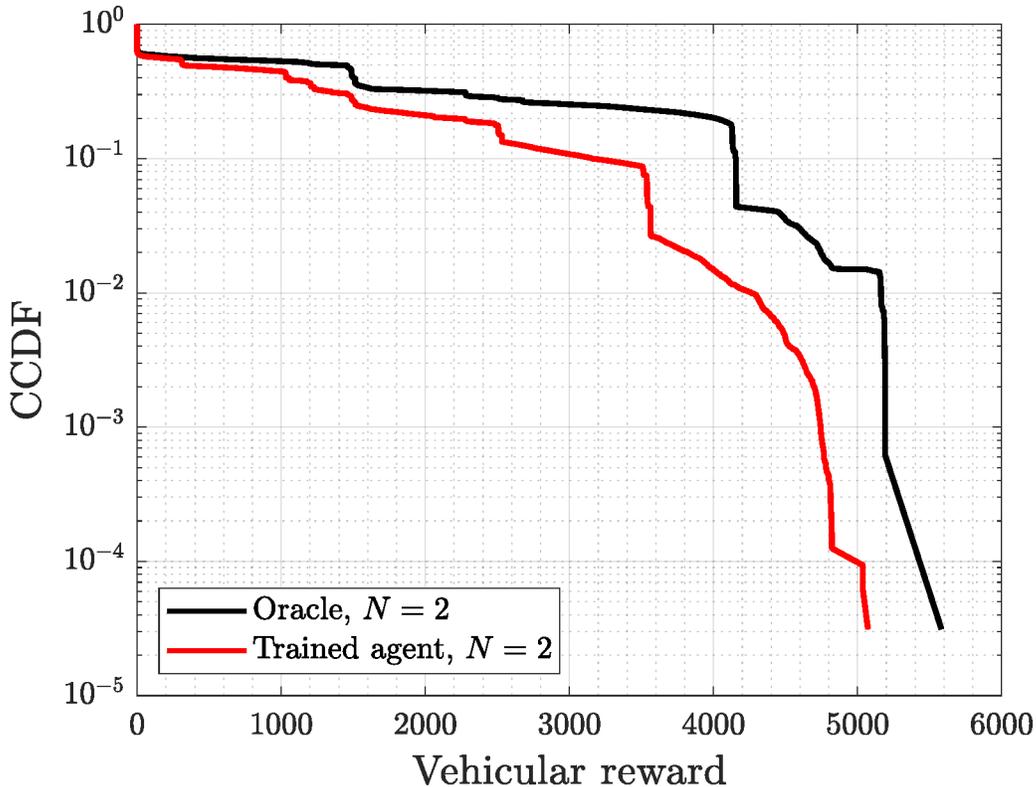}
\caption{The CCDF of the vehicular reward achieved by trained agents and oracle
for $N=2$.}
\label{fig:TrainedvsOracle}
\end{figure}

Finally, and in order to assess the quality of the learned cooperative perception policy, we consider a network of only $N=2$ vehicles\footnote{$N=2$ is selected to mitigate the effects of vehicular association and RB allocation, and to ensure the performance benefits are solely due to the aspects of cooperative perception.}, with
a maximum quadtree resolution level $L=2$. We compare our trained
agent to an oracle, which knows exactly the RoI's weights of each
vehicle and selects the quadtree blocks that maximizes the vehicular
satisfaction of both vehicles. Fig.\,\ref{fig:TrainedvsOracle} plots
the CCDF of the achieved vehicular rewards of all the vehicles in
both cases. It can be observed that the performance gap between the
trained agent and the oracle is small, which proves the effectiveness
of the proposed method in learning which quadtree blocks to transmit
in order to enhance the cooperative perception.

\section{Conclusion \label{sec:Conclusion}}

In this paper, we have studied the problem of associating vehicles,
allocating RBs and selecting the contents of CPMs in order to maximize
the vehicles' satisfaction in terms of the received sensory information
while considering the impact of the wireless communication. To solve
this problem, we have resorted to the DRL techniques where two RL
problems have been modeled. In order to overcome the huge action space
inherent to the formulation of our RL problems, we applied the dueling
and branching concepts. Moreover, we have proposed a federated RL
approach to enhance and accelerate the training process of the vehicles.
Simulation results show that policies achieving higher vehicular satisfaction
could be learned at both the RSU and vehicular sides leading to a
higher vehicular satisfaction.

\bibliographystyle{IEEEtran}
\bibliography{IEEEabrv,CooperativeSensing-Journal,CooperativePerception}

\end{document}